\pdfoutput=1
\documentclass[11pt]{article}
\usepackage[dvipsnames]{xcolor}

\usepackage[final]{acl}
\usepackage{multicol}
\usepackage{multirow}
\usepackage{amsmath,amsfonts}
\usepackage{algorithmic}
\usepackage{algorithm}
\usepackage{times}
\usepackage{latexsym, enumitem}
\usepackage{ragged2e}
\usepackage{booktabs}

\usepackage{pgfplots}
\usepackage{pgfplotstable}
\usepackage{tikz}
\pgfplotsset{compat=1.9}
\usepackage{pgf-pie}
\usetikzlibrary{patterns}
\usepackage{url}
\usepackage{hyperref}
\usepackage{makecell}
\usepackage{subcaption}
\usepackage{comment}

\usepackage[T1]{fontenc}
\usepackage[utf8]{inputenc}

\usepackage{microtype}

\usepackage{inconsolata}

\usepackage{graphicx}

\usepackage{amssymb}% http://ctan.org/pkg/amssymb
\usepackage{pifont}% http://ctan.org/pkg/pifont
\newcommand{\cmark}{\ding{51}}%
\newcommand{\xmark}{\ding{55}}%

\title{Compressing Long Context into Answer-Aligned Memory Embeddings for LLM Inference}

 \author{Md Mostafizer Rahman\textsuperscript{1}, Md Faizul Ibne Amin\textsuperscript{2}, Md Shahajada Mia\textsuperscript{2}, \\ \bf Yutaka Watanobe\textsuperscript{2},  Fang Liu\textsuperscript{1,3}\thanks{Corresponding author}\\
\textsuperscript{1}Lucy Family Institute for Data \& Society, University of Notre Dame, IN, USA \\
\textsuperscript{2}The University of Aizu, Aizuwakamatsu, Japan\\ \textsuperscript{3}Department of Applied and Computational Mathematics and Statistics, \\  University of Notre Dame, IN, USA \\
\texttt{\{mrahman3, fliu2\}@nd.edu}, \texttt{\{fiamin, d8262103, yutaka\}@u-aizu.ac.jp}}

\begin{document}
\maketitle
\begin{abstract}
Large language model (LLM) inference is constrained by the quadratic scaling of self-attention and the linear scaling of the KV cache, increasing latency, energy consumption, and GPU memory demand as context length scales.  Existing soft-compression methods either lack query-guided memory selection at inference time, train without answer-targeted supervision, or couple compression tightly to a specific decoder architecture. We propose a Context-to-Answer-Aligned Memory Compression (CMC) framework, which compresses long input contexts into compact Context Memory Embeddings (CMEs) aligned to any frozen decoder's embedding space, reducing inference costs without modifying decoder weights. CMC introduces a two-tier KV cache that combines question-guided CME selection with a local context window, and trains the compressor with answer-targeted distillation from a frozen LLM. Experiments across nine encoder-decoder combinations and four QA benchmarks show that CMC consistently outperforms the baseline, achieving up to 7.3 EM and 4.0 F1 point gains on SQuAD, while reducing inference time and energy consumption by up to 20\% and peak reserved GPU memory by up to 50\% at $3,000$ generation tokens. Ablation studies confirm that each architectural component and training objective contributes to the performance.
\end{abstract}
%Achieving efficient inference without sacrificing model performance remains a fundamental challenge in real-world LLM deployment. 
\section{Introduction}

\begin{figure*} [ht]
	\centering
		\includegraphics[width=1\linewidth]{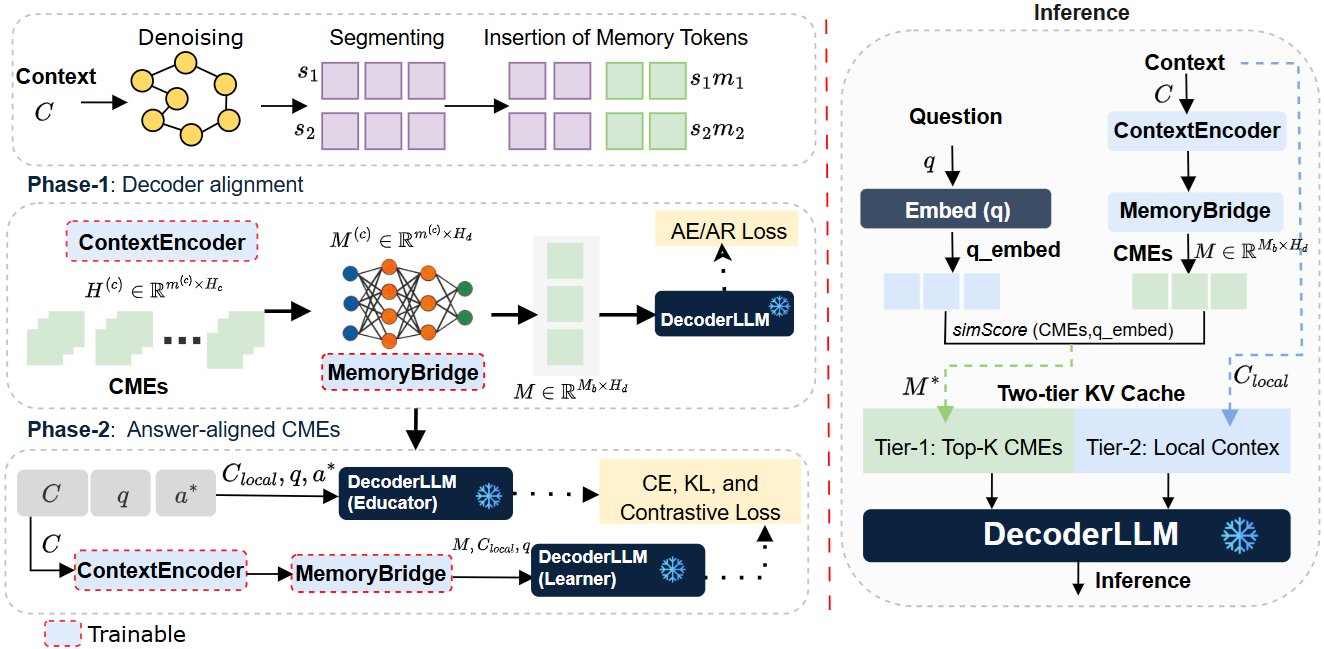}
	\caption{Overview of the CMC framework: ($i$)~context denoising and chunked compression via ContextEncoder, $ii$)~cross-architecture projection via MemoryBridge, and ($iii$)~two-tier KV cache inference via DecoderLLM with question-guided Tier-1 CME selection and Tier-2 local window.}
		\label{c2mframework}
\end{figure*}

Large language models (LLMs) have demonstrated exceptional performance across a wide range of natural language understanding and generation tasks~\cite{brown2020language,touvron2023llama2, jiang2023mistral,team2024gemma}. Central to their success is the Transformer architecture~\cite{vaswani2017attention}, whose self-attention mechanism enables rich contextual reasoning over input sequences. However, as the demand for processing longer documents, multi-turn dialogues, and multi-document reasoning grows, a fundamental bottleneck emerges: the computational cost of self-attention scales quadratically with sequence length, and the KV cache grows linearly, consuming substantial GPU memory, increasing inference latency, and driving up energy consumption with every additional token~\cite{pope2023efficiently,sheng2023flexgen}. These constraints become particularly acute in real-world deployments where contexts routinely span thousands of tokens, and throughput and energy efficiency are critical requirements alongside model quality.

Architectural methods address this bottleneck by redesigning the attention mechanism itself, either through sparse or local attention patterns~\cite{beltagy2020longformer,zaheer2020bigbird} or by extending positional encodings to accommodate longer sequences~\cite{chen2023extending,dong2024exploring}. While effective, these approaches require substantial architectural modification or continued pretraining, making them difficult to apply to already-deployed  LLMs. An alternative is hard-prompt compression, which discards tokens deemed redundant by self-information scoring or perplexity-based ranking~\cite{li2023compressing,pan-etal-2024-llmlingua}. Although model-agnostic, hard-prompt methods incur irreversible information loss and degrade on tasks requiring evidence synthesis across distant passages, such as multi-hop reasoning and long-document QA~\cite{li2025atacompressor}. In contrast, soft-compression methods encode the full context into a compact set of dense embeddings that a frozen decoder can attend to in place of the original tokens~\cite{chevalier2023adapting,ge2024incontext, mu2024learning,kim2024compressed,dai2025pcc}, achieving higher compression rates while preserving semantic content. \citet{dai2025pcc} propose PCC, a compressor-LLM framework that pretrains a lightweight encoder and converter to compress context into dense memory slots consumed by a frozen decoder. However, PCC supplies all compressed memory slots uniformly at inference time with no mechanism for query-guided selection, and trains solely on text reconstruction without answer-targeted supervision. More broadly, no existing soft-compression method simultaneously addresses cross-architecture deployment, answer-targeted training supervision, and query-guided memory selection at inference stage. 

To address these gaps, we propose a \textbf{Context-to-Answer-Aligned Memory Compression (CMC)} framework, illustrated in Figure~\ref{c2mframework}. CMC compresses long input contexts into compact Context Memory Embeddings (CMEs) projected into any frozen decoder's own embedding space, enabling efficient inference without modifying decoder weights. CMC comprises three key components: ($i$) a trainable \textbf{ContextEncoder} that applies graph-based context denoising and chunked compression to produce CMEs via fixed placeholder tokens; ($ii$) a \textbf{MemoryBridge}, a norm-calibrated two-layer MLP that projects CMEs into the frozen decoder's embedding space, enabling cross-architecture pairing of any encoder with any decoder; and ($iii$) a frozen \textbf{DecoderLLM} that generates responses conditioned on a two-tier KV cache combining top-$K$ CMEs selected by cosine similarity to the input question (Tier-1) and a local context window at full token resolution (Tier-2). The ContextEncoder and MemoryBridge are trained jointly in two phases: Phase-1 on autoencoding and autoregressive objectives to align CMEs with the decoder, and Phase-2 on knowledge distillation and contrastive memory-answer alignment using a frozen DecoderLLM to incorporate answer-targeted supervision. The main contributions of this work are as follows.

\begin{itemize}[leftmargin=12pt]
\item We propose CMC, a soft-compression framework   that incorporates graph-based context denoising to   improve CME quality and a two-tier KV cache   inference policy that combines question-guided   CME selection with a local   context window, providing \emph{both global} semantic   coverage \emph{and local} span-level precision without   full-context attention.

  \item We propose a two-phase training strategy. Phase-1 aligns CMEs with the frozen decoder via reconstruction objectives, and Phase-2 distills answer-targeted knowledge from a frozen DecoderLLM (educator) via cross-entropy, KL divergence, and contrastive memory-answer alignment losses, introducing answer-targeted training supervision for cross-architecture soft-compression.
%Mostafiz notes: To the best of my knowledge, the proposed two-phase training strategy — combining decoder-alignment objectives with answer-targeted distillation under a cross-architecture soft-compression framework — has not been explored in prior work. The literature review supports this, as existing methods either apply answer-targeted supervision within a fixed decoder architecture or perform cross-architecture compression without answer-targeted training, but none address both simultaneously. This is the basis for the "first" claim in the contribution bullet. That said, I recognize this is a strong claim and I want to make sure it is fully defensible before submission. If you feel the phrasing risks overstating the contribution or inviting unnecessary scrutiny from reviewers, I am happy to soften it to something like "the first cross-architecture soft-compression method with answer-targeted training supervision," which is more precisely scoped and easier to defend. Alternatively, I can remove the "first" qualifier entirely and let the experimental results speak for themselves.

\item We conduct comprehensive experiments across  nine encoder-decoder combinations and four QA   benchmarks (SQuAD, AdversarialQA, HotpotQA, CovidQA) and show that CMC consistently outperforms the baseline in EM and F1 with notable reductions in inference time, energy consumption, multiply–accumulate operation (MAC), and peak GPU memory. %\url{https://anonymous.4open.science/r/CMC-F840} (a public repository to be shared upon acceptance).
\end{itemize}

\section{Methods}

We propose the CMC framework to compress a long input context $C$ into compact CMEs that a frozen DecoderLLM attends to in place of the original tokens. Given context $C$ and question $q$, CMC learns a ContextEncoder $f_{\theta_c}$ and
MemoryBridge $g_{\phi}$ such that: 
\begin{align}
    \hat{a} & = f_{\theta_d}(\cdot \mid M^*,
    C_{\text{local}}, q),\mbox{ where}\\
    M^* &= \text{TopK}(g_{\phi}(
f_{\theta_c}(C)),\, q,\, K).
\end{align}
%Here $\hat{a}$ is the answer generated by the DecoderLLM, $M^* $ is the set of query-selected decoder-ready CMEs, and $C_{\text{local}}$ is the answer-centered local window of width $W$. $f_{\theta_c}$ and $g_{\phi}$ are the only trainable components, whereas $f_{\theta_d}$ remains frozen throughout. The following subsections detail context denoising and CME construction, cross-architecture projection, two-phase training, and two-tier KV cache inference policy, as illustrated in Figure~\ref{c2mframework}.
\noindent   $\hat{a}$ is the answer generated by the DecoderLLM $f_{\theta_d}$, $M^*$ is the set of  query-selected decoder-ready CMEs (formulated in Sections~\ref{sec:extraction} and \ref{sec:memorybridge}), and $C_{\text{local}}$ is the local window (formulated in Section~\ref{sec:c_local}). The ContextEncoder $f_{\theta_c}$ and MemoryBridge $g_{\phi}$ are the only trainable components, whereas $f_{\theta_d}$ remains frozen throughout. The following subsections detail each component as illustrated in Figure~\ref{c2mframework}.

\subsection{Context Denoising, Segmentation, and Memory Tokens}
%We apply graph-based context denoising %~\cite{rahman2026intuitivegraphllm,bugueno2023connecting} 
%to remove redundant tokens from the context before compression.   

We apply graph-based context denoising to remove low-salience tokens -- tokens that are semantically isolated from their immediate neighbors -- from the full context $C$ before
compression. Given context $C = (x_1, \ldots, x_n)$, we compute the cosine similarity between each token and its $g$  immediately downstream neighbors in embedding space. A token $x_i$ is retained if its cosine similarity to at least one of its $g$ subsequent neighbors meets or exceeds the tunable threshold $\tau$:
\begin{equation}
    C'_i = \left\{ x_i \in C \;\middle|\;
    \max_{j=i+1}^{\min(i+g,\,n)}
    \frac{\mathbf{e}_i \cdot \mathbf{e}_j}
    {\|\mathbf{e}_i\|\|\mathbf{e}_j\|}
    \geq \tau \right\}.
\end{equation}
$x_i$ is discarded if $C_i'$ is an empty set. For retained $x_i$, the denoised context $C'$ of $k\in(1,g]$ tokens is divided into $n_s$ non-overlapping segments, each of fixed length $t$, yielding $\mathcal{S} = \{s_1, s_2, \ldots, s_{n_s}\}$, where $n_s = \lceil k/t \rceil$. To each segment $s_i$, we append $m^{(c)}$ fixed placeholder tokens bounded by special \texttt{<MEM>} and \texttt{</MEM>} tokens, 
forming the augmented sequence $\tilde{s}_i$:
\begin{equation}\label{eq:tilde_s}
    \tilde{s}_i\!=\!\left[ s_i\Big\|
    \texttt{<MEM>} \Big\|
    \underbrace{\texttt{p}_1, \ldots,
    \texttt{p}_{m^{(c)}}}_{m^{(c)}
    \text{ placeholders}}
    \Big\|\texttt{</MEM>}
    \right],
\end{equation}
The number of placeholder tokens per segment is controlled by the compression rate $r = t / m^{(c)}$; a higher $r$ yields a more compact representation at the cost of reduced context coverage. We evaluate $r \in \{2, 4, 8\}$ in our experiments.

\subsection{CME Extraction with ContextEncoder}\label{sec:extraction}

Each augmented segment $\tilde{s}_i$ in Eq.~\eqref{eq:tilde_s} is processed independently by the ContextEncoder $f_{\theta_c}$, a trainable decoder-only causal language model. In particular, via causal self-attention, each of the $m^{(c)}$
placeholder tokens attends to all preceding chunk tokens in $s_i$, condensing the segment content into $m^{(c)}$ hidden state vectors $H^{(c)}_i\in \mathbb{R}^{m^{(c)} \times H_c}$ 
%The augmented sequence $\tilde{s}_i$ is defined with the following concatenation:
%\noindent where \texttt{<MEM>} and \texttt{</MEM>} are special boundary tokens added to the ContextEncoder vocabulary, and $\texttt{p}_1, \ldots, \texttt{p}_{m^{(c)}}$ are fixed placeholder tokens whose hidden states 
extracted as the segment-level CMEs from the final transformer layer:
\begin{align}
    H^{(c)}_i &= f_{\theta_c}(\tilde{s}_i)
    [\text{placeholder positions}],
\end{align}
\noindent where $H_c$ is the dimension of each hidden state vector. Though the placeholder tokens are fixed, their hidden states become informative CMEs through training the ContextEncoder weights. After
processing all $n_s$ segments, the full CME matrix is formed by concatenation:
\begin{equation}
    \mathbf{H}^{(c)} =
    [H^{(c)}_1;\, H^{(c)}_2;\, \ldots;\,
    H^{(c)}_{n_s}]
    \in \mathbb{R}^{M_b \times H_c},
\end{equation}
\noindent where $M_b = n_s \times m^{(c)}$ is the
total memory vectors representing the full context.

\begin{comment}

\subsection{Cross-Architecture Projection with
MemoryBridge}\label{sec:memorybridge}

The ContextEncoder and the frozen DecoderLLM may belong to different model families with different hidden dimensions $H_c$ and $H_d$. MemoryBridge is a trainable two-layer MLP that projects each CME vector $\mathbf{h} \in \mathbb{R}^{H_c}$ from  ContextEncoder into the DecoderLLM's embedding space $\mathbb{R}^{H_d}$.  Let $\hat{\mathbf{m}}$ denote the output of the two-layer MLP, 
we apply
\begin{equation}\label{eqn:norm}
    \tilde{\mathbf{m}} = \nu \cdot
    \tanh\!\left(
    \frac{\text{LayerNorm}(\hat{\mathbf{m}})}{\nu}
    \right),
\end{equation}
to prevent the projected CMEs from exceeding the range of the $\ell_2$ norm of the embeddings from DecoderLLM. $\nu = s \cdot \bar{\nu}$ in Eq.~\eqref{eqn:norm} is a norm cap, in which $s$ is a learnable scalar %, and $\bar{\nu}$ is the average $\ell_2$ norm of the frozen decoder's input embedding matrix over all tokens, computed once at initialization.
and $\bar{\nu}$ denotes the average $\ell_2$ norm of the frozen decoder's input token embedding vectors, computed across all vocabulary tokens once at initialization.

\end{comment}

\subsection{Cross-Architecture Projection with
MemoryBridge}\label{sec:memorybridge}
The ContextEncoder and the frozen DecoderLLM may belong to different model families with different hidden dimensions $H_c$ and $H_d$. MemoryBridge is a trainable two-layer MLP that projects each CME vector $\mathbf{h} \in \mathbb{R}^{H_c}$ from ContextEncoder into the DecoderLLM's embedding space $\mathbb{R}^{H_d}$. Let $\hat{\mathbf{m}}$ denote the output of the two-layer MLP after LayerNorm\footnote{See Appendix~\ref{app:hyperparameters} for the LayerNorm formulation used.}. We apply
\begin{equation}\label{eqn:norm}
    \tilde{\mathbf{m}} \!=\! \nu\tanh(\hat{\mathbf{m}})
    \cdot \min\!\left(1,\ \frac{2\nu}{\left\|\nu\tanh(\hat{\mathbf{m}})\right\|_2}\right),
\end{equation}
where $\tanh$ is applied coordinatewise and $\nu = s \cdot \bar{\nu}$, in which $s$ is a learnable scalar and $\bar{\nu}$ denotes the average $\ell_2$ norm of the frozen decoder's input token embedding vectors, computed across all vocabulary tokens once at initialization. The first factor, $\nu\,\tanh(\hat{\mathbf{m}})$, bounds each coordinate but not the $\ell_2$ norm of the vector; the second factor rescales the full vector whenever its $\ell_2$ norm exceeds $2\nu$, and leaves it unchanged otherwise. Eq.~\eqref{eqn:norm} guarantees $\|\tilde{\mathbf{m}}\|_2 \le 2\nu$, preventing the projected CMEs from exceeding twice the typical $\ell_2$ norm of the DecoderLLM's own token embeddings.

\begin{comment}
To prevent the projected CMEs from exceeding the typical magnitude of DecoderLLM embeddings, we apply
\begin{equation}
    \tilde{\mathbf{m}} = \nu \cdot
    \tanh\!\left(
    \frac{\text{LayerNorm}(\hat{\mathbf{m}})}{\nu}
    \right),
\end{equation}
\noindent where $\hat{\mathbf{m}}$ is the output of the two-layer MLP, and $\nu=s\cdot\bar{\nu}$ is a norm cap, with $s$ being a learnable scalar. Here, $\bar{\nu}$ denotes the average Euclidean ($\ell_2$) norm of the frozen decoder's input token embedding vectors, computed across all vocabulary tokens once at initialization.
\end{comment}
%averaged over all V token embedding vectors in the decoder’s input embedding matrix E ∈ RV ×Hd
%\noindent where $\hat{\mathbf{m}}$ is the output of the two-layer MLP, $\nu = s \cdot \bar{\nu}$ is a norm cap in which $s$ is a learnable scalar, and $\bar{\nu} = \frac{1}{V}\sum_{v=1}^{V} \|\mathbf{e}_v\|_2$ is the mean $\ell_2$ norm over all $V$ token embeddings of the frozen decoder, computed once at initialization.
%A final hard norm cap clips any vector whose norm exceeds $2\bar{\nu}$, ensuring the projected CMEs remain within the decoder's expected embedding range. 

Applying MemoryBridge $g_{\phi}$ to each segment's CME matrix $H^{(c)}_i$ yields its representation:
  $$ M^{(i)} = g_{\phi}(H^{(c)}_i)
    \in \mathbb{R}^{m^{(c)} \times H_d}.$$ 
After all $n_s$ segments are processed, the decoder-ready representations are concatenated to 
\begin{equation}\label{eq:M}
    M = [M^{(1)};\, M^{(2)};\, \ldots;\,
    M^{(n_s)}] \in \mathbb{R}^{M_b \times H_d}.
\end{equation}
to form the full memory matrix $M$ that provides the DecoderLLM with a compressed sequence of $H_d$-dimensional memory vectors in place of the original context tokens.

\subsection{Two-Phase Training}
%The ContextEncoder $f_{\theta_c}$ and MemoryBridge $g_{\phi}$ are trained jointly in two phases with the DecoderLLM $f_{\theta_d}$ frozen throughout. 
The ContextEncoder $f_{\theta_c}$ and MemoryBridge $g_\phi$ are the same trainable modules optimized continuously across both phases in sequence, with the DecoderLLM $f_{\theta_d}$ frozen throughout.

\paragraph{Phase-1: Decoder alignment.} Phase-1 aligns the CMEs with the decoder's embedding space via autoencoding (AE) and autoregressive (AR) objectives. The AE loss conditions the decoder on $M$ in Eq.~\eqref{eq:M}  to reconstruct the original context; the AR loss conditions the decoder on  the same $M$ to reconstruct only the first half of the context, leading to the loss function in Phase 1:
\begin{equation}\label{eqn:loss}
    \mathcal{L}_1 =
    \mathcal{L}_{\text{AE}} +\lambda_{\text{AR}}
    \mathcal{L}_{\text{AR}}.
\end{equation}
%Yes, there's overlap. Our AR loss, as implemented, conditions on the same memory M as the AE loss and computes a causal LM loss over only the first half of the context; it does not condition on the first half to predict the second half. Concretely, L_AE = −(1/T)Σ_{t=1}^{T} log P(c_t | M, c_<t) over the full context, and L_AR = −(1/T₁)Σ_{t=1}^{T₁} log P(c_t | M, c_<t) over just the first T₁ = T/2 tokens — the same functional form and the same M, evaluated over a shorter prefix. Every term in L_AR already appears in L_AE; the difference is only the normalization and which portion of the context is weighted. Both M = g_ϕ(f_θc(C)) — so f_θc and g_ϕ enter through M in both losses — and both losses use the frozen f_θd to compute the per-token log-probabilities.
%This loss formulation ensures that the CMEs carry sufficient information for the frozen decoder to recover the input.
This loss formulation ensures that CMEs carry sufficient information for the frozen decoder to reconstruct the context, where the AR term, with hyperparameter $\lambda_{\text{AR}}>0$, up-weights the reconstruction of the earlier portion relative to the full-context AE objective.

\paragraph{Phase-2: Answer-aligned CMEs.} Phase-2 fine-tunes the CMEs from Phase-1 toward answer-relevant content using the frozen DecoderLLM (educator) $f_{\theta_d}$, which receives the local window $C_{\text{local}}$ and question $q$. The same frozen weights $f_{\theta_d}$ are also invoked as the DecoderLLM (learner), conditioned on $M$ and $C_{\text{local}}$. The two DecoderLLMs share identical parameters and differ only in their input.

Three objectives act jointly: a cross-entropy loss $\mathcal{L}_{\text{CE}}$ supervises the DecoderLLM (learner) on gold answer tokens; a KL divergence loss $\mathcal{L}_{\text{KL}}$ aligns the DecoderLLM (learner)'s output distribution with the DecoderLLM (educator)'s; and a contrastive loss $\mathcal{L}_{\text{CL}}$~\citep{oord2018representation} pulls the
element-wise mean of the CMEs toward the DecoderLLM (educator)'s hidden state at the answer span. Taken together, the Phase-2 loss is:
\begin{equation}\label{eqn:l2}
    \mathcal{L}_2 =
    \mathcal{L}_{\text{CE}} +
    \lambda_{\text{KL}}\,\mathcal{L}_{\text{KL}} +
    \lambda_{\text{CL}}\,\mathcal{L}_{\text{CL}},
\end{equation}
\noindent where $\lambda_{\text{KL}}>0$ and $\lambda_{\text{CL}}>0$ are
hyperparameters. Our ablation study confirms that the three components in the Phase-2 loss in Eq.~\eqref{eqn:l2} act jointly -- removing any component degrades performance to the level of Phase-1 alone.

%a contrastive loss~\cite{oord2018representation} $\mathcal{L}_{\text{CL}}$ pulls the mean-pooled CME vector $\bar{M} = \frac{1}{M_b} \sum_{j=1}^{M_b} \tilde{\mathbf{m}}_j \in \mathbb{R}^{H_d}$ (the element-wise mean over all $M_b$ decoder-ready CME vectors) toward the educator's mean-pooled hidden state at the answer span.

\subsection{Inference with Two-Tier KV Cache
Policy}\label{sec:c_local}

At the inference stage, the frozen DecoderLLM generates responses conditioned on a two-tier KV cache built from the decoder-ready CME representation $M$ and question $q$. 

\paragraph{Tier-1: Question-guided top-$K$
selection.} The question tokens are embedded via the frozen decoder's embedding layer and mean-pooled to obtain a query vector $\mathbf{q}_{\text{vec}} \in \mathbb{R}^{H_d}$. Each CME $\tilde{\mathbf{m}}_j \in M$ is scored by cosine similarity:
\begin{equation}
    \text{score}_j =
    \frac{\tilde{\mathbf{m}}_j \cdot
    \mathbf{q}_{\text{vec}}}
    {\|\tilde{\mathbf{m}}_j\|
    \|\mathbf{q}_{\text{vec}}\|}.
\end{equation}
\noindent The top-$K$ CMEs are selected and re-ordered by original document position to form
$M^*$, where $K$ is a tunable hyperparameter controlling the breadth of semantic coverage. 

\paragraph{Tier-2: Local context window.} %A window of $W$ tokens is extracted from $C$ centered at character offset $\alpha$:
A $W$-token window $C_{\text{local}}$ is extracted from $C$ centered at the starting-token position  $\alpha$ of the gold answer span, subject to context boundary constraints:
\begin{align}
    C_{\text{local}} &= C[\text{start} : \text{start} + W],\mbox{ where}\\
\!\!\text{start} &= \max\left(\!0, \min\left(|C| \!-\! W,
    \alpha \!- W/2\right)\right).\notag
\end{align}
\noindent Tier-2 preserves the information critical region at full token resolution, providing the local textual precision that CMEs alone cannot supply. In our experiments, $\alpha$ is taken from benchmark annotations when available and otherwise computed by string matching of the information-critical region within the context $C$. Both CMC and the baseline use the same $\alpha$ in every comparison. In deployment settings where $\alpha$ is unavailable, a lightweight span localization step such as CME-guided or BM25 retrieval can be used; see Appendix~\ref{app:oracle_free}. 

\paragraph{KV cache assembly.} The prefill input is
\begin{equation}
    \text{input} = [\,M^* \;\|\;
    C_{\text{local}} \;\|\; q\,].
\end{equation}
\noindent %During autoregressive generation, the total KV cache length is bounded by $K + W$ tokens at all generation steps regardless of document length.%, enabling constant-memory inference over arbitrarily long contexts.
During autoregressive generation, the total KV cache length is bounded by $K + W + |q|$ tokens at all generation steps regardless of document length, where $|q|$ denotes the number of tokens in the question $q$.

\section{Experiments}

\subsection{Datasets}
%We evaluate CMC on four extractive QA benchmarks covering diverse reasoning types. SQuAD~\cite{rajpurkar2016squad} is a single-hop extractive QA dataset where answers are spans within a single passage. AdversarialQA~\cite{bartolo2020beat} extends this with adversarially constructed questions designed to challenge model comprehension. HotpotQA~\cite{yang2018hotpotqa} requires multi-hop reasoning across multiple supporting passages, while CovidQA~\cite{moller2020covid19} evaluates domain-specific QA on biomedical literature. We use two different sets of data for the training and validation. The small-sized set contains 10k training and 1k validation across datasets except CovidQA and the full-sized set contains full data. %For training and evaluation, we use 10,000 training samples and 1,000 test samples for SQuAD, AdversarialQA, and HotpotQA, and 1,500 training samples and 200 test samples for CovidQA due to its smaller size.
We evaluate CMC on four extractive QA benchmarks covering diverse reasoning types. SQuAD~\cite{rajpurkar2016squad} is a single-hop extractive QA dataset where answers are spans within a single passage. AdversarialQA~\cite{bartolo2020beat} extends this with adversarially constructed questions designed to challenge model comprehension. HotpotQA~\cite{yang2018hotpotqa} requires multi-hop reasoning across multiple supporting passages, while CovidQA~\cite{moller2020covid19} evaluates domain-specific QA on biomedical literature. We conduct experiments under two data settings. The \textbf{controlled-data setting} uses 10{,}000 training and 1{,}000 validation samples for SQuAD, AdversarialQA, and HotpotQA, and 1{,}500 training and 200 validation samples for CovidQA. The \textbf{full-data setting} uses the complete training and validation splits of SQuAD, AdversarialQA, and HotpotQA to enable direct comparison with published baselines; CovidQA is
evaluated under the small-data setting only due to its limited corpus size. This selection covers single-hop, adversarial, multi-hop, and
domain-specific reasoning, providing a broad evaluation of CMC across varied context lengths and answer types.

\begin{table*}[]
\centering

\small
\setlength{\tabcolsep}{5pt}
\begin{tabular}{l|l|cc|cc|cc|cc|cc}
\hline
& & \multicolumn{2}{c|}{\textbf{SQuAD}} &
\multicolumn{2}{c|}{\textbf{AdversarialQA}} &
\multicolumn{2}{c|}{\textbf{HotpotQA}} &
\multicolumn{2}{c|}{\textbf{CovidQA$^\dagger$}} &
\multicolumn{2}{c}{\textbf{Average}} \\
\textbf{DecoderLLM} & \textbf{Model} &
EM & F1 & EM & F1 & EM & F1 & EM & F1 & EM & F1 \\
\hline
\multicolumn{12}{c}{\textbf{Controlled-data setting}}
\\ \hline
\multirow{2}{*}{\shortstack[l]{Llama-3\\8B-Instruct}}
 & Llama (baseline)
   & 62.4 & 81.0
   & 34.3 & 53.3
   & 44.1 & 68.1
   & 22.5 & 63.8
   & 40.8 & 66.6 \\ 
 & CMC
   & \textbf{67.0} & \textbf{82.9}
   & \textbf{36.8} & \textbf{55.1}
   & \textbf{49.6} & \textbf{69.3}
   & \textbf{29.5} & \textbf{64.2}
   & \textbf{45.7} & \textbf{67.9} \\

\hline

%% ── Mistral-7B-Instruct-v0.3 ─────────────────────────────────
\multirow{2}{*}{\shortstack[l]{Mistral-7B\\Instruct-v0.3}}
 & Mistral (baseline)
   & \textbf{61.0} & \textbf{75.9}
   & 29.6 & 47.9
   & 53.0 & 71.1
   & 19.0 & 57.6
   & 40.7 & 63.1 \\
 & CMC
   & 60.2 & 75.5
   & \textbf{32.2} & \textbf{50.3}
   & \textbf{53.8} & \textbf{71.7}
   & \textbf{22.0} & \textbf{63.1}
   & \textbf{42.1} & \textbf{65.2} \\

\hline

%% ── Gemma-2-9B-IT ────────────────────────────────────────────
\multirow{2}{*}{\shortstack[l]{Gemma-2\\9B-IT}}
 & Gemma (baseline)
   & 55.1 & 78.1
   & 32.9 & 56.4
   & 40.1 & 67.0
   & 10.5& 62.1
   & 34.7 & 65.9 \\
 & CMC
   & \textbf{61.1} & \textbf{81.2}
   & \textbf{36.8} & \textbf{59.5}
   & \textbf{42.0} & \textbf{70.6}
   & \textbf{14.0} & \textbf{63.2}
   & \textbf{38.5} & \textbf{68.6} \\

\hline
\multicolumn{12}{c}{\textbf{Full-data setting}}\\
\hline
\multirow{2}{*}{\shortstack[l]{Llama-3\\8B-Instruct}}
 & Llama (baseline)
   & 48.28 & 74.27
   & 32.47 & 52.55
   & 44.31 & 67.28 
   
   & - & -
   & 41.69 & 64.70 \\ 
 & CMC
   & \textbf{55.56} & \textbf{78.29}
   & \textbf{35.43} & \textbf{53.42}
   & \textbf{47.77} & \textbf{67.83}
   
   & \textbf{-} & \textbf{-}
   & \textbf{46.25} & \textbf{66.51} \\

\hline

%% ── Mistral-7B-Instruct-v0.3 ─────────────────────────────────
\multirow{2}{*}{\shortstack[l]{Mistral-7B\\Instruct-v0.3}}
 & Mistral (baseline)
   & \textbf{45.36} & \textbf{69.26}
  
   & 26.37 & 45.91
    & \textbf{53.41} & \textbf{71.13}
   & - & -
   & \textbf{41.71} & \textbf{62.10} \\
 & CMC
   & 30.67 & 55.16
   & \textbf{29.77} & \textbf{49.26}
   & {48.33} & {65.49}
   
   & - & -
   & {35.37} & {55.38} \\

\hline

%% ── Gemma-2-9B-IT ────────────────────────────────────────────
\multirow{2}{*}{\shortstack[l]{Gemma-2\\9B-IT}}
 & Gemma (baseline)
   & 40.91 & 71.68
   & 29.77 & 53.17
   & 37.80 & 65.60
   
   & -& -
   & 36.16 & 63.48\\
 & CMC
   & \textbf{44.13} & \textbf{73.27}
   & \textbf{33.33} & \textbf{56.24}
   & \textbf{41.90} & \textbf{69.10}
   
   & - & -
   & \textbf{39.79} & \textbf{66.20} \\

\hline

%\multicolumn{12}{c}{NB: Due to the small data size of the CovidQA dataset falls under the Controlled-data settings.}\\

\end{tabular}
\caption{EM and F1 on four QA benchmarks under controlled-data and full-data settings. \textbf{Bold} denotes the higher score per row. $^\dagger$CovidQA is evaluated under the
controlled-data setting only due to its limited corpus size.}
\label{tab:main_results}

\end{table*}

\subsection{Experimental Settings}

All experiments are conducted on NVIDIA
A100 80\,GB GPU. The DecoderLLM is loaded in 4-bit
NF4 quantisation with double quantisation enabled
and kept frozen throughout training and inference.
We evaluate three ContextEncoders -- GPT2-Large,
OPT-1.3B, and OPT-2.7B -- paired with three
frozen DecoderLLMs -- Llama-3-8B-Instruct,
Mistral-7B-Instruct-v0.3, and Gemma-2-9B-IT --
yielding nine encoder-decoder combinations. The
ContextEncoder and MemoryBridge are trained jointly for one epoch using AdamW
(learning rate $=1\times10^{-5}$, $\beta=(0.9,\,0.98)$). Full hyperparameter details are provided in Appendix~\ref{app:hyperparameters}.

\paragraph{Evaluation Metrics.}
We evaluate CMC on task performance and inference efficiency. For task performance, we report Exact Match (EM) and F1. For efficiency, we measure inference time decomposed into compression, prefill, and decode stages, energy consumption in kWh via CodeCarbon~\cite{courty2024codecarbon}, MACs, and peak allocated and reserved GPU memory. Efficiency metrics are evaluated on SQuAD using GPT2-Large, Mistral, and Llama  at $r=4$, across generation token budgets $T \in \{512, 1024, 2048, 3000\}$.

\subsection{Baselines}

%We evaluate CMC against two categories of baselines. \noindent\textbf{Full-context baseline.} The primary baseline passes the answer-centered context window directly to the frozen DecoderLLM without any compression, using the gold answer offset to centre the window. We evaluate this across three DecoderLLMs --- Llama-3-8B-Instruct, Mistral-7B-Instruct-v0.3, and Gemma-2-9B-IT --- using identical training data, evaluation splits, and inference settings as CMC, ensuring a fair and controlled comparison. \noindent\textbf{Published baselines.} For full-size dataset experiments on SQuAD, HotpotQA, and AdversarialQA, we compare against LLMLingua-2~\cite{pan-etal-2024-llmlingua}, AutoCompressor~\cite{chevalier2023adapting}, xRAG~\cite{cheng2024xrag}, ICAE~\cite{ge2024incontext}, PCC-lite~\cite{dai2025pcc}, and PCC-large~\cite{dai2025pcc}. 

We evaluate CMC against two groups of baselines. The \noindent\textbf{primary baseline} passes the local context window directly to the frozen DecoderLLM without compression, using the gold answer offset to center the window. We run this baseline across all three DecoderLLMs using identical training data, evaluation splits, and inference settings as CMC, ensuring a controlled comparison. %This baseline represents the performance upper bound and efficiency lower bound for each decoder configuration.
The second baseline type is the \textbf{published baselines.} For full-data experiments on SQuAD, HotpotQA, and AdversarialQA, we compare against one hard-prompt method -- LLMLingua-2~\cite{pan-etal-2024-llmlingua} -- and four soft-compression methods -- AutoCompressor~\cite{chevalier2023adapting}, 
xRAG~\cite{cheng2024xrag}, ICAE~\cite{ge2024incontext}, and PCC-lite and PCC-large~\cite{dai2025pcc}. %Published results are taken from \cite{dai2025pcc}. Since these methods use different decoder architectures, comparisons are indicative rather than strictly controlled.

\subsection{Results}

\subsubsection{Task Performance} \label{task_performance}
Table~\ref{tab:main_results} reports EM and F1 under both data  settings at $r=4$; full results across all nine encoder-decoder combinations and $r \in \{2,4,8\}$ are in Appendix~\ref{app:detailed_results}. \noindent\textbf{Controlled-data setting.} CMC outperforms the baselines on the majority of datasets. With Llama, CMC achieves the largest gains, improving EM by 4.6, 2.5, 5.5, and 7.0 points on SQuAD, AdversarialQA, HotpotQA, and CovidQA, respectively (average $+$4.9 EM). For Mistral, CMC improves on three of four datasets for a net average gain of 1.4 EM. For Gemma, CMC improves consistently across all four datasets, with the largest F1 gains on AdversarialQA ($+$3.1) and HotpotQA ($+$3.6). %\noindent\textbf{Full-data setting.} Llama and Gemma continue to benefit, with Llama gaining 7.3 EM on SQuAD (48.3$\rightarrow$55.6) and Gemma 3.2 EM (40.9$\rightarrow$44.1). Mistral is an exception, underperforming its baseline by 14.7 EM on SQuAD (45.4$\rightarrow$30.7) and regressing on HotpotQA (53.4$\rightarrow$48.3), which we attribute to its sensitivity to compression artefacts under the more variable context length distribution of the full-data split.
\noindent\textbf{Full-data setting.} CMC continues to benefit Llama and Gemma, with Llama gaining 7.28 EM on SQuAD (48.28$\rightarrow$55.56) and Gemma gaining 3.22 EM (40.91$\rightarrow$44.13). Mistral shows a mixed pattern.  CMC improves over the Mistral baseline on AdversarialQA (26.37$\rightarrow$29.77, +3.40 EM) but underperforms the Mistral baseline by 14.7 EM on SQuAD (45.4$\rightarrow$30.7) and regresses on HotpotQA (53.4$\rightarrow$48.3). %, which we attribute to Mistral's sensitivity to compression artefacts under the more variable context length distribution of the full-data split. 
 All three decoders share an identical Phase-2 configuration, including a fixed KD step budget that does not scale with training set size, while Phase-1 steps scale directly with it ($\sim$8.8$\times$ more steps on full-data vs.\ controlled-data SQuAD); we attribute Mistral's SQuAD/HotpotQA regressions to this training imbalance, to which Llama and Gemma appear less sensitive under the identical settings.

%For the fair comparison with the reference baseline models across datasets, we have used EM and F1 scores of CMC with Llama-3-8B-Instruct, as shown in Table \ref{tab:main_comparison}. CMC obtained superior performance on both metrics across three SQuAD, HotPotQA, and Adversarial datasets only exception for EM score on SQuAD dataset. 

%Table~\ref{tab:main_comparison} compares CMC against published baselines on full-data setting of SQuAD, HotpotQA, and AdversarialQA using Llama. Since published baselines use different decoder architectures \cite{dai2025pcc}, these comparisons are indicative rather than strictly controlled. CMC achieves the best F1 on all three datasets, surpassing PCC-Large on SQuAD F1 (78.29), HotpotQA F1 (67.83), and AdversarialQA F1 (53.42). On HotpotQA, CMC also leads on EM (47.77 for PCC-Large), a 7.8 point gain, which is particularly notable given that HotpotQA requires multi-hop reasoning across multiple passages. On the other hand, CMC underperforms PCC-Large on EM for SQuAD (55.56 vs 60.04) and AdversarialQA (35.43 vs 39.37), likely reflecting the difference in decoder capacity since PCC uses a larger decoder architecture. AutoCompressor, xRAG, and ICAE perform substantially below compared to CMC.%, consistent with their reported results in .

\begin{table}[ht]
\centering
\small
\setlength{\tabcolsep}{3pt}
\begin{tabular}{l|cc|cc|cc}
\hline
\textbf{Method}
& \multicolumn{2}{c|}{\textbf{SQuAD}}
& \multicolumn{2}{c|}{\textbf{HotpotQA}}
& \multicolumn{2}{c}{\textbf{Adv.QA}} \\
& EM & F1 & EM & F1 & EM & F1 \\
\hline
AutoCompressor & 0.35  & 21.46 & 0.29  & 16.29 & 2.00  & 14.09 \\
xRAG           & 3.46  & 18.19 & 16.29 & 27.51 & 3.47  & 13.75 \\
ICAE           & 21.63 & 45.69 & 26.68 & 35.16 & 11.70 & 27.98 \\
LLMLingua-2    & 32.18 & 51.20 & 44.18 & 55.72 & 24.80 & 35.41 \\
PCC (Lite, 4$\times$)  & 57.44 & 75.83
                       & 42.20 & 50.37
                       & 37.83 & 50.36 \\
PCC (Large, 4$\times$) & \textbf{60.04} & 77.76
                       & 39.97 & 48.19
                       & \textbf{39.37} & 52.56 \\ 

\textbf{CMC}           & {55.56} & \textbf{78.29}
                       & \textbf{47.77} & \textbf{67.83}
                       & {35.43} & \textbf{53.42} \\ \hline

\end{tabular}
%\caption{Comparison against published baselines on full-data setting. The best EM and F1 results with CMC are presented.}
\caption{Comparison of CMC with published baselines under the full-data setting on SQuAD, HotpotQA, and AdversarialQA. The best EM and F1 scores for each dataset are highlighted in bold.}
\label{tab:main_comparison}
\end{table}

%\footnotetext{On HotpotQA, CMC underperforms the full-context baseline (EM=53.41, F1=71.13), suggesting that multi-hop reasoning remains challenging under compression. See Appendix~\ref{app:detailed_results} for full baseline comparisons.}

Table~\ref{tab:main_comparison} compares CMC against published baselines on the full-data setting. Although PCC-lite shares the same GPT2-Large encoder as CMC, all published baselines use different decoder architectures~\cite{dai2025pcc}. % these comparisons are therefore indicative rather than strictly controlled.
%Table~\ref{tab:main_comparison} compares CMC against published baselines on the full-data setting using Llama as the DecoderLLM. Since published baselines use different decoder architectures~\cite{dai2025pcc}, these comparisons are indicative rather than strictly controlled. 
CMC achieves the best F1 on all three datasets, surpassing PCC-Large by 0.53 F1 on SQuAD, 19.64 on HotpotQA, and 0.86 on AdversarialQA. On HotpotQA, CMC also leads on EM (47.77 vs 39.97 for PCC-Large), a 7.8-point gain particularly notable given the multi-hop reasoning requirement. CMC underperforms PCC-Large on EM for SQuAD (55.56 vs 60.04) and AdversarialQA (35.43 vs 39.37), likely reflecting PCC's use of a larger encoder/decoder architecture. AutoCompressor and xRAG achieve low EM on SQuAD, HotpotQA, and AdversarialQA, while ICAE and LLMLingua-2 perform substantially below CMC across all three datasets.  A qualitative case study illustrating CMC behavior across four outcome categories is provided in Appendix~\ref{app:case_study}.

%The optimal number of CMEs generation is important for the accurate inference in QA task. In CMC,  we selected three different compression rates $r\in\{2, 4, 8\}$ for CMEs generation. Figure \ref{comrate_parameter_sensitivity} shows the performance (EM and F1) of CMC across  these compression rates, DecoderLLMs, and ContextEncoders. When we use $r=2$, it allows more CMEs generation that impact negetively on EM and F1 performances regarless the DecoderLLMs and ContextEncoders. In contrary, with $r=8$ reduces the CMEs generation and the fewer number of CMEs couldn't properly represent the context information which also negatively impact the CMC performance. While we use the $r=4$, it ensures the better performance (EM and F1) across the DecoderLLMs and ContextEncers. THus the optimal number of CMEs that properly represent the context helped boost performance.  

\begin{figure*}[ht]
%\vspace{-3mm}
 \captionsetup[subfigure]{font=scriptsize}
     \centering
     \begin{subfigure}[b]{.16\linewidth}
         \captionsetup{justification=centering}
         \begin{tikzpicture}[scale=1]
            \begin{axis}[
    ybar=.05cm,
    every node near coord/.append style={font=\scriptsize},
    legend style={font=\scriptsize},
    tick label style={font=\scriptsize},
    ylabel near ticks, ylabel shift={-6pt},
    %xlabel shift={-10pt},
    %width=\textwidth,
    width=3.4cm,
    height=3cm,
    every node near coord/.append style={
                        anchor=west,
                        rotate=75
                },
    enlargelimits=.28,
    enlarge y limits={0.1,upper},
    legend style={at={(3,1.45)},
    anchor=north, legend columns=3},
    legend style={draw=none},
    ymin=64, 
    ylabel={EM(\%)},
    symbolic x coords={r=2, r=4, r=8},
    xtick=data,
    %nodes near coords,
    ytick={60, 62, 64, 65, 66, 67,  68,70},
    %x tick label style={rotate=25,anchor=east},
    grid=both,
    %nodes near coords,
    nodes near coords align={vertical},
    bar width=3.2pt,
    %ymajorgrids=true,
    label style={font=\scriptsize},
    ]
\addplot [draw=RedOrange , semithick, pattern=crosshatch dots, pattern color = RedOrange ] coordinates {(r=2,65.7) (r=4,66.8) (r=8,66.7)};  % Macro F1-Score for 0.001

\addplot [draw=RedOrange, semithick, pattern=north west lines,  pattern color = RedOrange] coordinates {(r=2,65.76) (r=4,66.8) (r=8,66.7)};  % Macro F1-Score for 0.005

\addplot [draw=RedOrange, semithick, pattern=horizontal lines,  pattern color = RedOrange] coordinates {(r=2,65.3) (r=4,67.0) (r=8,66.7)}; % Macro F1-Score for 0.01

%\addplot [draw=black, semithick, pattern=north east lines, pattern color = black] coordinates {($Sorting$,26326) ($Searching$,26173) };

\legend{OPT-2.7B, OPT-1.3B, GPT2-Large}
\end{axis}
 
        \end{tikzpicture}
        \caption{CMC (Llama): EM}
         \label{roberta-bilstm-twitter2}
     \end{subfigure}
     \hfill   
     \begin{subfigure}[b]{0.16\linewidth}
        \captionsetup{justification=centering}
         \begin{tikzpicture}[scale=1]
            \begin{axis}[
    ybar=.05cm,
    every node near coord/.append style={font=\scriptsize},
    legend style={font=\scriptsize},
    tick label style={font=\scriptsize},
    ylabel near ticks, ylabel shift={-6pt},
    %xlabel shift={-10pt},
    %width=\textwidth,
    width=3.4cm,
    height=3cm,
    every node near coord/.append style={
                        anchor=west,
                        rotate=75
                },
    enlargelimits=.28,
    enlarge y limits={0.1,upper},
    legend style={at={(0.5,-0.32)},
    anchor=north, legend columns=2},
    ymin=80, 
    ylabel={F1 (\%)},
    symbolic x coords={r=2, r=4, r=8},
    xtick=data,
    %nodes near coords,
    ytick={80,81, 82, 83,84, 85, 86, 87,88, 89,90},
    %x tick label style={rotate=25,anchor=east},
    grid=both,
    %nodes near coords,
    nodes near coords align={vertical},
    bar width=3.2pt,
    %ymajorgrids=true,
    label style={font=\scriptsize},
    ]
\addplot [draw=NavyBlue, semithick, pattern=crosshatch dots, pattern color = NavyBlue] coordinates {(r=2,81.13) (r=4,82.83) (r=8,82.74)};  % Macro F1-Score for 0.001

\addplot [draw=NavyBlue, semithick, pattern=north west lines,  pattern color = NavyBlue] coordinates {(r=2,81.13) (r=4,82.83) (r=8,82.74)};  % Macro F1-Score for 0.005

\addplot [draw=NavyBlue, semithick, pattern=horizontal lines,  pattern color = NavyBlue] coordinates {(r=2,81.07) (r=4,82.86) (r=8,82.74)}; % Macro F1-Score for 0.01

%\addplot [draw=black, semithick, pattern=north east lines, pattern color = black] coordinates {($Sorting$,26326) ($Searching$,26173) };

%\legend{OPT-2.7B, OPT-1.3B, GPT2-Large}
\end{axis}
 
        \end{tikzpicture}
        \caption{CMC (Llama): F1}
         \label{roberta-gru-sentiment1401}
     \end{subfigure}
         \hfill   
     \begin{subfigure}[b]{0.16\linewidth}
        \captionsetup{justification=centering}
         \begin{tikzpicture}[scale=1]
            \begin{axis}[
    ybar=.05cm,
    every node near coord/.append style={font=\scriptsize},
    legend style={font=\scriptsize},
    tick label style={font=\scriptsize},
    ylabel near ticks, ylabel shift={-6pt},
    %xlabel shift={-10pt},
    %width=\textwidth,
    width=3.4cm,
    height=3cm,
    every node near coord/.append style={
                        anchor=west,
                        rotate=75
                },
    enlargelimits=.28,
    enlarge y limits={0.1,upper},
    legend style={at={(0.92,1.45)},
    legend style={draw=none},
    anchor=north, legend columns=-1},
    ymin=56, 
    ylabel={EM(\%)},
    symbolic x coords={r=2, r=4, r=8},
    xtick=data,
    %nodes near coords,
    ytick={50, 52, 54, 56, 57, 58, 59, 60, 61},
    %x tick label style={rotate=25,anchor=east},
    grid=both,
    %nodes near coords,
    nodes near coords align={vertical},
    bar width=3.2pt,
    %ymajorgrids=true,
    label style={font=\scriptsize},
    ]
\addplot [draw=RedOrange, semithick, pattern=crosshatch dots, pattern color = RedOrange] coordinates {(r=2,57.4) (r=4,59.9) (r=8,60.0)};  % Macro F1-Score for 0.001

\addplot [draw=RedOrange, semithick, pattern=north west lines,  pattern color = RedOrange] coordinates {(r=2,57.4) (r=4,59.9) (r=8,60.0)};  % Macro F1-Score for 0.005

\addplot [draw=RedOrange, semithick, pattern=horizontal lines,  pattern color = RedOrange] coordinates {(r=2,57.4) (r=4,60.02) (r=8,60.0)}; % Macro F1-Score for 0.01

%\addplot [draw=black, semithick, pattern=north east lines, pattern color = black] coordinates {($Sorting$,26326) ($Searching$,26173) };

%\legend{OPT-2.7B, OPT-1.3B, GPT2-Large}
\end{axis}
 
        \end{tikzpicture}
        \caption{CMC (Mistral): EM}
         \label{roberta-gru-sentiment1402}
     \end{subfigure}
    \hfill   
     \begin{subfigure}[b]{0.16\linewidth}
        \captionsetup{justification=centering}
         \begin{tikzpicture}[scale=1]
            \begin{axis}[
    ybar=.05cm,
    every node near coord/.append style={font=\scriptsize},
    legend style={font=\scriptsize},
    tick label style={font=\scriptsize},
    ylabel near ticks, ylabel shift={-6pt},
    %xlabel shift={-10pt},
    %width=\textwidth,
    width=3.4cm,
    height=3cm,
    every node near coord/.append style={
                        anchor=west,
                        rotate=75
                },
    enlargelimits=.28,
    enlarge y limits={0.1,upper},
    legend style={at={(1.3,1.45)},
    legend style={draw=none},
    anchor=north, legend columns=-1},
    ymin=72, 
    ylabel={F1 (\%)},
   symbolic x coords={r=2, r=4, r=8},
    xtick=data,
    %nodes near coords,
    ytick={70,71,72,73,74,75,76,77,78,79,80},
    %x tick label style={rotate=25,anchor=east},
    grid=both,
    %nodes near coords,
    nodes near coords align={vertical},
    bar width=3.2pt,
    %ymajorgrids=true,
    label style={font=\scriptsize},
    ]
\addplot [draw=NavyBlue, semithick, pattern=crosshatch dots, pattern color = NavyBlue] coordinates {(r=2,73.91) (r=4,75.33) (r=8,75.32)};  % Macro F1-Score for 0.001

\addplot [draw=NavyBlue, semithick, pattern=north west lines,  pattern color = NavyBlue] coordinates {(r=2,73.91) (r=4,75.33) (r=8,75.32)};  % Macro F1-Score for 0.005

\addplot [draw=NavyBlue, semithick, pattern=horizontal lines,  pattern color = NavyBlue] coordinates {(r=2,73.90) (r=4,75.51) (r=8,75.32)}; % Macro F1-Score for 0.01

%\addplot [draw=black, semithick, pattern=north east lines, pattern color = black] coordinates {($Sorting$,26326) ($Searching$,26173) };

\legend{OPT-2.7B, OPT-1.3B, GPT2-Large}
\end{axis}
 
        \end{tikzpicture}
        \caption{CMC (Mistral): F1}
         \label{roberta-gru-sentiment1403}
     \end{subfigure}
    \hfill   
     \begin{subfigure}[b]{0.16\linewidth}
        \captionsetup{justification=centering}
         \begin{tikzpicture}[scale=1]
            \begin{axis}[
    ybar=.05cm,
    every node near coord/.append style={font=\scriptsize},
    legend style={font=\tiny},
    tick label style={font=\scriptsize},
    ylabel near ticks, ylabel shift={-6pt},
    %xlabel shift={-10pt},
    %width=\textwidth,
    width=3.4cm,
    height=3cm,
    every node near coord/.append style={
                        anchor=west,
                        rotate=75
                },
    enlargelimits=.28,
    enlarge y limits={0.1,upper},
    legend style={at={(0.5,-0.32)},
    anchor=north, legend columns=2},
    ymin=60, 
    ylabel={EM(\%)},
    symbolic x coords={r=2, r=4, r=8},
    xtick=data,
    %nodes near coords,
    ytick={60, 61,62,63,64,65},
    %x tick label style={rotate=25,anchor=east},
    grid=both,
    %nodes near coords,
    nodes near coords align={vertical},
    bar width=3.2pt,
    %ymajorgrids=true,
    label style={font=\scriptsize},
    ]
\addplot [draw=RedOrange, semithick, pattern=crosshatch dots, pattern color = RedOrange] coordinates {(r=2,60.9) (r=4,61.0) (r=8,61.1)};  % Macro F1-Score for 0.001

\addplot [draw=RedOrange, semithick, pattern=north west lines,  pattern color = RedOrange] coordinates {(r=2,60.9) (r=4,61) (r=8,61.0)};  % Macro F1-Score for 0.005

\addplot [draw=RedOrange, semithick, pattern=horizontal lines,  pattern color = RedOrange] coordinates {(r=2,60.9) (r=4,61) (r=8,61.0)}; % Macro F1-Score for 0.01

%\addplot [draw=black, semithick, pattern=north east lines, pattern color = black] coordinates {($Sorting$,26326) ($Searching$,26173) };

%\legend{OPT-2.7B, OPT-1.3B, GPT2-Large}
\end{axis}
 
        \end{tikzpicture}
        \caption{CMC (Gemma): EM}
         \label{roberta-gru-sentiment1404}
     \end{subfigure}
         \hfill   
     \begin{subfigure}[b]{0.16\linewidth}
     \centering
        \captionsetup{justification=centering}
         \begin{tikzpicture}[scale=1]
            \begin{axis}[
    ybar=.05cm,
    every node near coord/.append style={font=\scriptsize},
    legend style={font=\tiny},
    tick label style={font=\scriptsize},
    ylabel near ticks, ylabel shift={-6pt},
    %xlabel shift={-10pt},
    %width=\textwidth,
    width=3.4cm,
    height=3cm,
    every node near coord/.append style={
                        anchor=west,
                        rotate=75
                },
    enlargelimits=.28,
    enlarge y limits={0.1,upper},
    legend style={at={(0.5,-0.32)},
    anchor=north, legend columns=-1},
    ymin=80, 
    ylabel={F1 (\%)},
    symbolic x coords={r=2, r=4, r=8},
    xtick=data,
    %nodes near coords,
    ytick={80,81,82,83,84},
    %x tick label style={rotate=25,anchor=east},
    grid=both,
    %nodes near coords,
    nodes near coords align={vertical},
    bar width=3.2pt,
    %ymajorgrids=true,
    label style={font=\scriptsize},
    ]
\addplot [draw=NavyBlue, semithick, pattern=crosshatch dots, pattern color = NavyBlue] coordinates {(r=2,81.02) (r=4,81.15) (r=8,81.18)};  % Macro F1-Score for 0.001

\addplot [draw=NavyBlue, semithick, pattern=north west lines,  pattern color = NavyBlue] coordinates {(r=2,81.02) (r=4,81.15) (r=8,81.15)};  % Macro F1-Score for 0.005

\addplot [draw=NavyBlue, semithick, pattern=horizontal lines,  pattern color = NavyBlue] coordinates {(r=2,81.02) (r=4,81.15) (r=8,81.15)}; % Macro F1-Score for 0.01

%\addplot [draw=black, semithick, pattern=north east lines, pattern color = black] coordinates {($Sorting$,26326) ($Searching$,26173) };

%\legend{OPT-2.7B, OPT-1.3B, GPT2-Large}
\end{axis}
 
        \end{tikzpicture}
        \caption{CMC (Gemma): F1}
         \label{roberta-gru-sentiment1405}
     \end{subfigure}
     %\vspace{-2mm}     
          \caption{EM and F1 scores on SQuAD (controlled-data setting) across compression rates $r \in \{2, 4, 8\}$ for three DecoderLLMs paired with three ContextEncoders.}
        \label{comrate_parameter_sensitivity}
        %\vspace{-6mm}
        
\end{figure*}
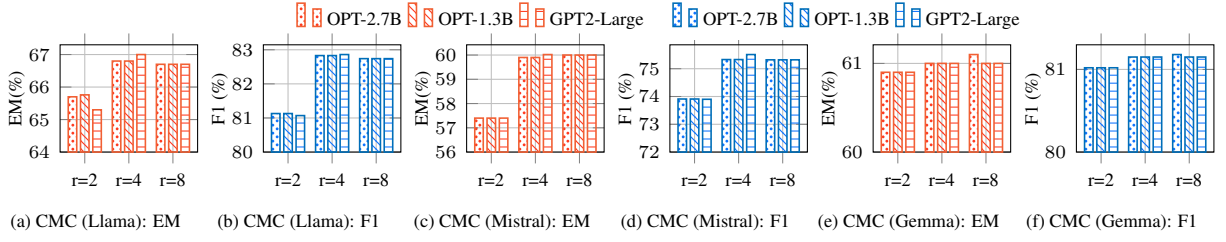

\subsubsection{Generation and Use of CMEs}
The number of CMEs generated per context chunk is controlled by the compression rate $r$, where a smaller $r$ produces more CMEs per chunk and a larger $r$ produces fewer. Figure~\ref{comrate_parameter_sensitivity} shows the EM and F1 sensitivity to $r \in \{2, 4, 8\}$ across all nine encoder-decoder combinations on SQuAD. $r=2$ consistently yields the lowest performance. The larger CME count introduces noise into the Tier-1 prefix, diluting answer-relevant signals and making it harder for the frozen DecoderLLMs to locate the correct span. In contrast, $r=8$ reduces the CME count to the point where important contextual information is lost, particularly for longer passages. The intermediate setting $r=4$ achieves the best or tied-best EM in 6 of 9 combinations. However, the optimal rate is mildly decoder-dependent: for example, Llama strongly favors $r=4$ across all encoders (EM gains of 1.1--1.7 points over $r=8$), while Mistral and Gemma show near-flat performance between $r=4$ and $r=8$ (differences $\leq 0.1$ EM), suggesting decoder-specific sensitivity to compression granularity. %We perform a case study with baseline and CMC in Table \ref{tab:case_study} in Appendix.

\subsubsection{Computational Efficiency}\label{CMC_efficiency}

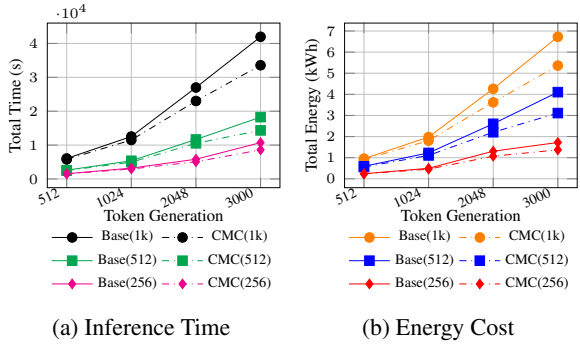
\begin{figure}[ht]
%\vspace{-2mm}
     \centering
    
       \begin{subfigure}[b]{0.49\linewidth}      
       \centering
        \captionsetup{justification=centering}
         \begin{tikzpicture}[scale=.9]
            \begin{axis}[
    ylabel={Total Time (s)},
    xlabel={Token Generation},
    ylabel near ticks, ylabel shift={-5pt},
    xlabel shift={-8pt},
    width=\linewidth,
    width=5cm,
    height=4cm,
    grid,
    grid style={gray!50},
    grid=both,
    %ymode=log,                         % <-- use log scale here
    %log basis y={10},                 % base-10 log scale
    %ymin=100, ymax=2000,             % adjust range for log scale
    %ymax=132,
    %xmin=0, xmax=13,
    %ymin=20, ymax=75,
    % xtick={0,20,40,60,80,100},
    ytick={0,10000,20000,30000,40000,50000},
    symbolic x coords={512,1024, 2048,3000},
    xtick=data,
    x tick label style={rotate=25,anchor=east},
    ymajorgrids=true,
    grid style=solid,
    label style={font=\scriptsize},
    every node near coord/.append style={font=\tiny},
    legend style={font=\tiny},
    tick label style={font=\tiny},
    legend style={at={(0.5,-0.2)},
    legend style={draw=none},
    anchor=north, legend columns=2},
]

%\addplot [mark=*, red, dashed] table[x=N,y=No_Sym] {data.dat};

\addplot[mark=otimes*, black, solid, every mark/.append style={solid, fill=black}]
    coordinates {
(512,6043.28)
(1024,12529.52)
(2048,26985.80)
(3000,41963.20)
    };
    \addlegendentry{Base(1k)}
    
\addplot[mark=otimes*, black, dashdotted, every mark/.append style={solid, fill=black}]
    coordinates {
(512,5763.45)
(1024,11504.01)
(2048,23012.03)
(3000,33561.58)
    };
    \addlegendentry{CMC(1k)}

\addplot[mark=square*, Green, solid, every mark/.append style={solid, fill=Green}]
    coordinates {
(512,2591.49)
(1024,5371.51)
(2048,11654.82)
(3000,18250.79)
    };
\addlegendentry{Base(512)}

\addplot[mark=square*, Green, dashdotted, every mark/.append style={solid, fill=Green}]
    coordinates {
(512,2474.97)
(1024,4956.70)
(2048,10522.99)
(3000,14340.18)

    };
\addlegendentry{CMC(512)}

\addplot[mark=diamond*, magenta, solid, every mark/.append style={solid, fill=magenta}]
    coordinates {
(512,1543.38)
(1024,3201.91)
(2048,5822.62)
(3000,10724.70)
    };
\addlegendentry{Base(256)}

\addplot[mark=diamond*, magenta, dashdotted, every mark/.append style={solid, fill=magenta}]
    coordinates {
(512,1477.71)
(1024,2943.98)
(2048,5080.43)
(3000,8589.51)
    };
\addlegendentry{CMC(256)}

\end{axis}
        \end{tikzpicture}
        %\vspace{-5mm}
        \caption{Inference Time}
         \label{llama_inference}
     \end{subfigure}
          \hfill
         \begin{subfigure}[b]{0.49\linewidth}
     \centering
         \captionsetup{justification=centering}
         \begin{tikzpicture}[scale=.9]
            \begin{axis}[
    ylabel={Total Energy (kWh)},
    xlabel={Token Generation},
    ylabel near ticks, ylabel shift={-5pt},
    xlabel shift={-8pt},
    width=\linewidth,
    width=5cm,
    height=4cm,
    grid,
    grid style={gray!50},
    grid=both,
    %ymode=log,                         % <-- use log scale here
    %log basis y={10},                 % base-10 log scale
    %ymin=100, ymax=2000,             % adjust range for log scale
    %ymax=132,
    %xmin=0, xmax=13,
    %ymin=20, ymax=75,
    % xtick={0,20,40,60,80,100},
    ytick={0,1.0,2.0,3.0,4.0,5.0,6.0,7.0},
    symbolic x coords={512,1024, 2048,3000},
    xtick=data,
    x tick label style={rotate=25,anchor=east},
    ymajorgrids=true,
    grid style=solid,
    label style={font=\scriptsize},
    every node near coord/.append style={font=\tiny},
    legend style={font=\tiny},
    tick label style={font=\tiny},
    legend style={at={(0.5,-0.2)},
    legend style={draw=none},
    anchor=north, legend columns=2},
]

%\addplot [mark=*, red, dashed] table[x=N,y=No_Sym] {data.dat};

\addplot[mark=otimes*, orange, solid, every mark/.append style={solid, fill=orange}]
    coordinates {
(512,0.9564)
(1024,1.9773)
(2048,4.2625)
(3000,6.7254)
    };
    \addlegendentry{Base(1k)}
    
\addplot[mark=otimes*, orange, dashdotted, every mark/.append style={solid, fill=orange}]
    coordinates {
(512,0.9125)
(1024,1.8092)
(2048,3.6208)
(3000,5.3623)
    };
    \addlegendentry{CMC(1k)}

\addplot[mark=square*, blue, solid, every mark/.append style={solid, fill=blue}]
    coordinates {
(512,0.5938)
(1024,1.2267)
(2048,2.6056)
(3000,4.1036)
    };
\addlegendentry{Base(512)}

\addplot[mark=square*, blue, dashdotted, every mark/.append style={solid, fill=blue}]
    coordinates {
(512,0.5562)
(1024,1.1082)
(2048,2.2056)
(3000,3.1128)

    };
\addlegendentry{CMC(512)}

\addplot[mark=diamond*, red, solid, every mark/.append style={solid, fill=red}]
    coordinates {
(512,0.2404)
(1024,0.4978)
(2048,1.3079)
(3000,1.7200)
    };
\addlegendentry{Base(256)}

\addplot[mark=diamond*, red, dashdotted, every mark/.append style={solid, fill=red}]
    coordinates {
(512,0.2298)
(1024,0.4576)
(2048,1.0709)
(3000,1.3718)
    };
\addlegendentry{CMC(256)}

\end{axis}
        \end{tikzpicture}
         %\vspace{-5mm}
        \caption{Energy Cost}
         \label{llama_energy}
     \end{subfigure}    
       %\begin{subfigure}[b]{0.49\linewidth}      \centering
        %\captionsetup{justification=centering}
         %\begin{tikzpicture}[scale=.9]
          %  \input{energy_mistral_squad}
        %\end{tikzpicture}
        %\vspace{-5mm}
        %\caption{Mistral — Energy Cost}
         %\label{time_curve}
     %\end{subfigure}
%\caption{Stage-wise inference time and total energy consumption for generating $T \in \{512, 1024, 2048, 3000\}$ tokens on SQuAD across three sample sizes (256, 512, and 1{,}000), using Llama-3-8B-Instruct as the DecoderLLM, GPT2-Large as the ContextEncoder, $r=4$, and batch size 8. Energy is measured via CodeCarbon~\cite{courty2024codecarbon}.}
        %\label{llama_inference_energy} 
\caption{Inference time and energy consumption of CMC vs.\ the baseline on SQuAD across $T \in \{512, 1024, 2048, 3000\}$ tokens. Parenthetical labels denote sample size (256, 512, 1{,}000).}

%\caption{Inference time and energy consumption of CMC and Llama on SQuAD across generation token budgets $T \in \{512, 1024, 2048, 3000\}$ and three evaluation set sizes. CMC/Base(.) denotes data size.}
\label{llama_inference_energy}
        %\vspace{-5mm}
\end{figure}

Figure~\ref{llama_inference_energy} shows inference time and energy cost for $T \in \{512, 1024, 2048, 3000\}$ tokens on SQuAD for three evaluation sample sizes. CMC consistently reduces both metrics relative to the baseline across all token budgets and sample sizes. At 1{,}000 samples, the total time reduction ranges from 4.6\% at $T=512$ (6{,}043\,s $\rightarrow$ 5{,}763\,s) to 20.0\% at $T=3{,}000$ (41{,}963\,s $\rightarrow$ 33{,}562\,s), with corresponding energy reductions of 4.6\% and 20.3\%, respectively.  This scaling effect arises because CMC's compression overhead is approximately constant ($\approx$25\,s regardless of $T$), while decode savings grow with generation length due to the bounded two-tier KV cache. The same trend holds across all sample sizes, confirming the gains reflect the two-tier KV cache architecture rather than dataset-specific effects. Stage-wise inference time and energy breakdowns for Llama and Mistral are provided in Appendix \ref{app:detailed_results_efficiency}. %Tables~\ref{tab:llama_time}--\ref{tab:mistral_energy} in the appendix. %Stage-wise breakdowns and peak GPU memory results for Llama and Mistral are provided in Tables~\ref{tab:llama_time}--\ref{tab:mistral_memory} in the appendix.
%The near-identical time and energy reductions arise from CodeCarbon’s energy apportionment methodology. CodeCarbon measures the total energy consumed during an experiment via hardware power monitoring, then apportions stage-wise energy proportionally to wall-clock time (as noted in the caption of Tables 11 and 13). Under this methodology, stage energy ≈ P · t where P is the GPU power draw and t is the stage duration. Since the GPU operates at near- constant utilisation during inference (decode stage is fully GPU-bound), P is approximately constant, making energy reduction ≈ time reduction by construction.  This proportionality is confirmed by the data: the ratio of energy reduction to time reduction is consistently ≈ 1.0 across all token budgets (0.99 at T = 512, 1.04 at T = 1024, 1.02 at T = 2048, 1.01 at T = 3000). The small deviations from 1.0 reflect minor variations in GPU power draw across stages.

\begin{figure}[ht]
%\vspace{-3mm}
     \centering
     \begin{subfigure}[b]{0.49\linewidth}
         \captionsetup{justification=centering}
         \begin{tikzpicture}[scale=1]
            \begin{axis}[
    ybar=.1cm,
    every node near coord/.append style={font=\tiny},
    legend style={font=\tiny},
    tick label style={font=\tiny},
    ylabel near ticks, ylabel shift={-6pt},
    %xlabel shift={-10pt},
    %width=\textwidth,
    width=4.8cm,
    height=3.2cm,
    every node near coord/.append style={
                        anchor=west,
                        rotate=75
                },
    enlargelimits=.25,
    enlarge y limits={0.1,upper},
    legend style={at={(0.5,-0.42)},
    anchor=north, legend columns=-1},
    ymin=2, 
    ylabel={Peak allocated (GB)},
    xlabel={Token Generation},
    xlabel shift={-5pt},
    symbolic x coords={512,1024,2048,3000},
    xtick=data,
    %nodes near coords,
    ytick={2, 4, 6, 8, 10, 12},
    %x tick label style={rotate=25,anchor=east},
    grid=both,
    %nodes near coords,
    nodes near coords align={vertical},
    bar width=5pt,
    %ymajorgrids=true,
    label style={font=\tiny},
    legend style={draw=none}
    ]
\addplot [draw=magenta, semithick, pattern=crosshatch dots, pattern color = magenta] coordinates {(512,10.017) (1024,10.017) (2048,10.017) (3000,10.159)};  % Macro F1-Score for 0.001

%\addplot [draw=blue, semithick, pattern=north west lines,  pattern color = blue] coordinates {($h=128$,78.01) ($h=256$,80.74) ($h=512$,80.33)};  % Macro F1-Score for 0.005

\addplot [draw=teal, semithick, pattern=horizontal lines,  pattern color = teal] coordinates {(512,8.853) (1024,8.853) (2048,8.853) (3000,8.853)}; % Macro F1-Score for 0.01

%\addplot [draw=black, semithick, pattern=north east lines, pattern color = black] coordinates {($Sorting$,26326) ($Searching$,26173) };

\legend{Baseline, CMC}
\end{axis}
 
        \end{tikzpicture}
        \caption{Peak allocated memory}
         \label{roberta-bilstm-twitter}
     \end{subfigure}
     \hfill   
     \begin{subfigure}[b]{0.49\linewidth}
        \captionsetup{justification=centering}
         \begin{tikzpicture}[scale=1]
            \begin{axis}[
    ybar=.1cm,
    every node near coord/.append style={font=\tiny},
    legend style={font=\tiny},
    tick label style={font=\tiny},
    ylabel near ticks, ylabel shift={-6pt},
    %xlabel shift={-10pt},
    %width=\textwidth,
    width=4.8cm,
    height=3.2cm,
    every node near coord/.append style={
                        anchor=west,
                        rotate=75
                },
    enlargelimits=.25,
    enlarge y limits={0.1,upper},
    legend style={at={(0.5,-0.42)},
    anchor=north, legend columns=-1},
    ymin=2, 
    ylabel={Peak reserved (GB)},
    xlabel={Token Generation},
    xlabel shift={-5pt},
    symbolic x coords={512,1024,2048,3000},
    xtick=data,
    %nodes near coords,
    ytick={2, 6, 10, 14,18,22},
    %x tick label style={rotate=25,anchor=east},
    grid=both,
    %nodes near coords,
    nodes near coords align={vertical},
    bar width=5pt,
    %ymajorgrids=true,
    label style={font=\tiny},
    legend style={draw=none},
    ]
\addplot [draw=red, semithick, pattern=crosshatch dots, pattern color = red] coordinates {(512,11.197 ) (1024,11.197 ) (2048,11.199 ) (3000,19.689)};  % Macro F1-Score for 0.001

%\addplot [draw=blue, semithick, pattern=north west lines,  pattern color = blue] coordinates {($h=128$,78.01) ($h=256$,80.74) ($h=512$,80.33)};  % Macro F1-Score for 0.005

\addplot [draw=green, semithick, pattern=horizontal lines,  pattern color = green] coordinates {(512,9.836) (1024,9.836) (2048,9.836) (3000,9.836)}; % Macro F1-Score for 0.01

%\addplot [draw=black, semithick, pattern=north east lines, pattern color = black] coordinates {($Sorting$,26326) ($Searching$,26173) };

\legend{Baseline, CMC}
\end{axis}
 
        \end{tikzpicture}
        \caption{Peak reserved memory}
         \label{roberta-gru-sentiment140}
     \end{subfigure}
     %\vspace{-2mm}
\caption{Peak allocated and peak reserved GPU memory (GB) for CMC vs.\ the baseline on 1{,}000 SQuAD samples across $T \in \{512, 1024, 2048, 3000\}$ tokens.}
\label{parameter_sensitivity}
        %\vspace{-6mm}
\end{figure}
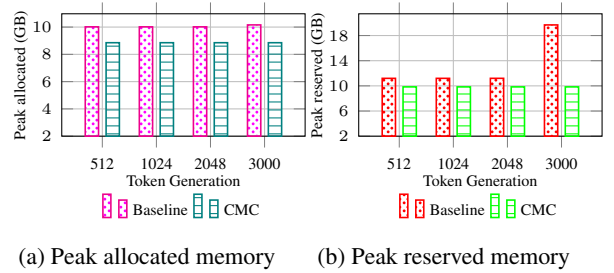

Figure~\ref{parameter_sensitivity} shows peak allocated and peak reserved GPU memory for Llama
across $T \in \{512, 1024, 2048, 3000\}$ on SQuAD
(1{,}000 samples, GPT2-Large, $r=4$). Peak allocated memory remains nearly constant for CMC ($\approx$8.85\,GB) while the baseline increases from 10.0\,GB at $T=512$ to 10.2\,GB at $T=3{,}000$, yielding a reduction of $\approx$1.2\,GB.  Peak reserved memory reveals a more
substantial difference: the baseline spikes to
19.7\,GB at $T=3{,}000$ while CMC maintains a
near-constant 9.8\,GB, corresponding to a 50\% reduction. Mistral
results in the appendix show an even larger effect
(24.3\,GB $\rightarrow$ 9.1\,GB, a 62.5\% reduction). These
reductions arise because baseline
inference accumulates KV cache entries
proportionally to $T$, whereas CMC's two-tier
policy caps the cache at a fixed budget. Detailed memory results for both decoders are provided in Appendix~\ref{app:memory_appendix}. Analytical attention MAC analysis  in Appendix~\ref{app:macs_appendix} confirms that these savings are architectural rather than hardware-specific.

%Tables~\ref{tab:llama_memory}--\ref{tab:mistral_memory} in appendix.

\section{Ablation Study}

Table~\ref{tab:ablation} presents the ablation results
on SQuAD. \textbf{Architectural ablations (A1--A3):}
Removing graph-based denoising (A1) reduces EM by
26.6 points relative to CMC, 22.0 points
below the baseline, confirming that denoising is essential for CME quality rather than merely a computational optimization. Random
CME selection (A2) reduces EM by 8.1 points: the
same number of CME slots reach the decoder but are decorrelated from the question, reducing Tier-1 prefix informativeness. Removing Tier-2 (A3) produces the largest drop ($-$50.0 EM). The near-identical AE/AR losses between A3 and CMC (1.40/0.73 vs 1.38/0.70) confirm the deficit arises at inference rather than from degraded CME quality, establishing Tier-2 as structurally indispensable. 
\textbf{Phase-2 training objective ablations (A4--A6).} All converge to EM~$=$~0.589 -- identical to Phase-1 only (A6). Removing either $\mathcal{L}_{\text{KL}}$ or $\mathcal{L}_{\text{CL}}$ individually is as damaging as removing Phase-2 entirely, revealing that all three objectives must act jointly: $\mathcal{L}_{\text{CE}}$ provides token-level supervision, $\mathcal{L}_{\text{KL}}$ aligns the learner with the educator's distribution, and $\mathcal{L}_{\text{CL}}$ grounds the CME prefix in the answer embedding space. %Sensitivity analyses of top-$K$ CME selection, local window size, denoising threshold, and tier-2 window localisation are provided in Appendix~\ref{app:topk}, \ref{app:window}, \ref{app:tau}, and \ref{app:oracle_free}.
%% ============================================================
%% Ablation Study Table + Description
%% Dataset: SQuAD (10k train / 1k val)
%% Decoder: Llama-3-8B-Instruct | Encoder: GPT2-Large | r=4
%% ============================================================
%% ── TABLE ──────────────────────────────────────────────────
\begin{table}[!htb]
\centering
\small
\setlength{\tabcolsep}{2pt}
\begin{tabular}{p{3.9cm}|cccc}
\hline
\textbf{Model} & \textbf{EM} & \textbf{F1}
               & $\boldsymbol{\Delta}$\textbf{EM}
               & $\boldsymbol{\Delta}$\textbf{F1} \\ \hline
Llama (baseline) & 0.624 & 0.810 & $-$0.046 & $-$0.019 \\
\hline
\textbf{CMC}   & \textbf{0.670} & \textbf{0.829}
                      & --- & --- \\
\hline
\textit{Architecture ablations}& & & &  \\
~w/o graph denoising (A1)     & 0.404 & 0.564 & $-$0.266 & $-$0.265 \\
~w/o query-guided top-$K$ (A2) & 0.589 & 0.757 & $-$0.081 & $-$0.072 \\
~w/o Tier-2 local window (A3) & 0.170 & 0.247 & $-$0.500 & $-$0.582 \\
\hline
\textit{Training objective ablations}& & & & \\
~w/o $\mathcal{L}_{\text{CL}}$ (A4) & 0.589 & 0.757 & $-$0.081 & $-$0.072 \\
~w/o $\mathcal{L}_{\text{KL}}$ (A5) & 0.589 & 0.757 & $-$0.081 & $-$0.072 \\
~Phase-1 only, no Phase-2 (A6)     & 0.589 & 0.757 & $-$0.081 & $-$0.072 \\
\hline
\end{tabular}
\caption{Ablation study on SQuAD (controlled-data setting, GPT2-Large, $r=4$, Llama). $\Delta$EM and $\Delta$F1 are relative to CMC.}
\label{tab:ablation}
\end{table}

\textbf{Sensitivity analyses} of local window size and denoising threshold are provided in Appendices \ref{app:window} and~\ref{app:tau}, respectively. A \textbf{deployment analysis} of Tier-2 window localization without gold annotations is provided in Appendix~\ref{app:oracle_free}.

\section{Related Work}

\noindent\textbf{Architectural methods.}
Sparse and local attention mechanisms~\cite{beltagy2020longformer,zaheer2020bigbird} reduce quadratic complexity by restricting each token's receptive field, while positional interpolation~\cite{chen2023extending},
YaRN~\cite{peng2023yarn}, and LongLoRA~\cite{chen2023longlora} extend the context window of pretrained LLMs through rotary embedding modifications. FlashAttention~\cite{dao2022flashattention} reduces memory bandwidth costs without approximation, and KV cache management strategies such as H2O~\cite{zhang2023h2o}, and
StreamingLLM~\cite{xiao2024efficient} evict low-importance entries during decoding. These
methods require architectural modification, continued pretraining, or operate only on
already-computed caches without reducing prefill cost, and none produce a compact input
representation consumable by an unmodified frozen LLM.

\noindent\textbf{Hard-prompt compression.}
Selective Context~\cite{li2023compressing} prunes tokens by
self-information, LLMLingua~\cite{jiang2023llmlingua} and
LongLLMLingua~\cite{jiang2023longllmlingua} apply perplexity-based
budget control, LLMLingua-2~\cite{pan-etal-2024-llmlingua} distils
compression decisions into a lightweight classifier, and
QGC~\cite{cao2024retaining} weights tokens by query relevance. These methods are
model-agnostic but incur irreversible information
loss, as discarded tokens cannot be recovered by
the decoder, making them susceptible to performance
degradation on tasks requiring cross-passage
evidence synthesis such as multi-hop reasoning.

\noindent\textbf{Soft-compression.}
Soft-compression methods encode the full context into dense embeddings consumed by a frozen decoder, avoiding hard token removal. RMT~\cite{bulatov2022recurrent} propagates memory tokens recurrently across segments. AutoCompressor~\cite{chevalier2023adapting} and ICAE~\cite{ge2024incontext} compress long documents into soft summary vectors via autoencoding objectives. Gisting~\cite{mu2024learning} compresses instructions into gist tokens, and CCM~\cite{kim2024compressed} targets online interaction scenarios by compressing accumulating KV pairs via a conditional LoRA adapter. 500xCompressor \cite{500xcompressor} compresses the entire context into a small set of learned tokens, and then uses the KV representation of these tokens for the downstream QA task. In RAG settings, xRAG~\cite{cheng2024xrag} and COCOM~\cite{rau2024cocom} compress retrieved passages into dense representations for a single specific target decoder. \citet{dai2025pcc} introduce a decoupled compressor-LLM framework with cross-architecture projection but without query-guided memory selection or answer-targeted training supervision. ATACompressor~\cite{li2025atacompressor} introduces query-guided selective encoding but does not address cross-architecture deployment or
answer-targeted memory alignment. No existing method simultaneously addresses all three gaps,
which CMC is designed to fill.

\section{Conclusion}
\vspace{-.08cm}
We presented CMC, a soft-compression framework that compresses long contexts into compact CMEs for computation- and energy- efficient inference with frozen LLMs. Across nine encoder-decoder combinations and four extractive QA benchmarks, CMC consistently outperforms the baseline in the majority of configurations under the controlled-data setting, and achieves the best F1 against published soft-compression baselines under full-data conditions. At $T=3{,}000$ tokens, CMC reduces inference time and energy by 20\% and peak reserved GPU memory by up to 62.5\%, enabling deployment at generation lengths that would otherwise exhaust GPU memory. Ablation studies confirm that graph-based denoising, the Tier-2 local window, and all three Phase-2 objectives are each necessary; removing any single component causes substantial performance degradation.

\section*{Limitations}
%\vspace{-1.5mm}
While CMC demonstrates consistent improvements in EM, F1, and inference efficiency, we acknowledge the following limitations.
%\begin{itemize}[leftmargin=12pt]
    %\item The Tier-2 local window relies on the gold answer character offset $\alpha$, which is   available in annotated QA benchmarks but   unavailable at deployment time. A lightweight   span localisation step would be required for real-world use.
%\item 
First, all experiments are conducted on extractive QA benchmarks; the effectiveness of CMC on abstractive QA, summarisation, and RAG has not been evaluated and may require adjustments to the answer-targeted distillation objective. Moreover, CMC is trained and evaluated within each benchmark independently; cross-dataset generalization (e.g., training on SQuAD and evaluating on HotpotQA) has not been evaluated.
%\item 
Second, CMC uses a fixed compression rate $r \in  \{2, 4, 8\}$, which may over-compress short passages or under-compress long ones; adaptive rate scheduling conditioned on context length is left for future work.
%\item 
Lastly, CMC is evaluated on English-language QA benchmarks only; its effectiveness on non-English or multilingual contexts, where tokenisation and semantic similarity properties differ substantially, has not been investigated.

%\end{itemize}

\section*{Ethics Statement}
No ethical approval was needed for this study.

\section*{Availability Statement}
Code and additional resources for this study are available at  \url{https://github.com/mostafiz26/CMC}.

\bibliography{custom}

%\clearpage

\appendix

\section{Appendix}

\subsection{Hyperparameters}
\label{app:hyperparameters}

Table~\ref{tab:hyperparams} reports the full set of
hyperparameters used across all experiments. %All results are reported from a single training run due to the high computational cost of training with ContextEncoders ranging from 774M to 2.7B parameters paired with frozen DecoderLLMs ranging from 7B to 9B parameters, across large-scale QA datasets of up to 90{,}000 training examples. 
The ContextEncoder and MemoryBridge are optimised jointly; the frozen DecoderLLM receives no gradient updates at any stage. The two-tier KV cache hyperparameters $K$ and $W$ are tuned per dataset and encoder-decoder configuration within the ranges reported below.

\begin{table}[H]
\centering
\small
\setlength{\tabcolsep}{4pt}
\begin{tabular}{lp{3.8cm}}
\toprule
\textbf{Hyperparameter} & \textbf{Value} \\
\midrule
\multicolumn{2}{l}{\textit{Hardware}} \\
GPU & NVIDIA A100 80\,GB \\
\midrule
\multicolumn{2}{l}{\textit{DecoderLLM}} \\
Quantisation & 4-bit NF4, double quantisation \\
Compute dtype & bfloat16 \\
Decoder training & Frozen (no gradient updates) \\
\midrule
\multicolumn{2}{l}{\textit{Optimiser (ContextEncoder + MemoryBridge)}} \\
Optimiser & AdamW \\
Learning rate & $1 \times 10^{-5}$ \\
Weight decay & $0.01$ \\
$\beta$ & $(0.9,\ 0.98)$ \\
Gradient clip norm & $0.6$ \\
Training epochs & $1$ \\
Batch size & $\{4, 8\}$ (tuned per config) \\
\midrule
\multicolumn{2}{l}{\textit{ContextEncoder}} \\
Architectures & GPT2-Large, OPT-1.3B, OPT-2.7B \\
Max input length & 3{,}072 tokens \\
\midrule
\multicolumn{2}{l}{\textit{MemoryBridge}} \\
Architecture & Two-layer MLP \\
Normalisation & LayerNorm ($\epsilon=10^{-5}$) \\
Activation & GELU \\
Output activation & Norm-calibrated tanh \\
\midrule
\multicolumn{2}{l}{\textit{Graph-based token denoising}} \\
Cosine similarity threshold $\tau$ & $0.05$ \\
Neighbourhood window $k$ & $3$ \\
Minimum keep interval & Every 16 tokens \\
IG warmup steps & $200$ \\
\midrule
\multicolumn{2}{l}{\textit{Chunking and compression}} \\
Chunk size $t$ & $256$ tokens \\
Compression rates $r$ & $\{2, 4, 8\}$ \\
Memory token cap per chunk & $128$ \\
Min memory tokens per chunk & $4$ \\
Warmup memory tokens & $8$ \\
\midrule
\multicolumn{2}{l}{\textit{Two-tier KV cache}} \\
Tier-1 top-$K$ CMEs & $K \in \{30, 40\}$ \\
Tier-2 local window $W$ & $W \in \{130, 256\}$ tokens \\
\midrule
\multicolumn{2}{l}{\textit{Phase-1 AE/AR}} \\
%AE loss weight $\lambda_{\text{AE}}$  &  1.0\\
AR loss weight $\lambda_{\text{AR}}$  &  1.0\\ \midrule
\multicolumn{2}{l}{\textit{Phase-2 QA distillation}} \\
KD steps per epoch & $800$ \\
KD batch size & $2$ \\
CE loss weight $\lambda_{\text{CE}}$ & $1.0$ \\
KL loss weight $\lambda_{\text{KL}}$ & $0.5$ \\
Contrastive loss weight $\lambda_{\text{CL}}$ & $0.1$ \\
Contrastive temperature $\delta$ & $0.07$ \\
Local window drop probability & $0.5$ \\
Local window chars (educator) & $1{,}200 - 2,000$ \\
Local window tokens (student) & $130$ \\
\midrule
\multicolumn{2}{l}{\textit{Inference}} \\
Max new tokens & $32$ \\
Decoder max length & $1{,}536$ tokens \\
\bottomrule
\end{tabular}
\caption{Hyperparameters used in all CMC experiments.}
\label{tab:hyperparams}
\end{table}

\paragraph{LayerNorm details.}\label{applayernorm} All LayerNorm operations in MemoryBridge use PyTorch's
default \texttt{nn.LayerNorm}: $\text{LayerNorm}(\mathbf{x}) = \boldsymbol{\gamma} \odot
(\mathbf{x} - \mu)/\sqrt{\sigma^2 + \epsilon} + \boldsymbol{\beta}$, where $\mu$ and $\sigma^2$ are the mean and variance of $\mathbf{x}$ over the feature dimension, $\epsilon = 10^{-5}$ (Table~\ref{tab:hyperparams}), and $\boldsymbol{\gamma}, \boldsymbol{\beta}$ are learnable per-dimension scale and shift parameters trained jointly with the rest of MemoryBridge.

%% ============================================================
%% TABLE 1: SQuAD
%% ============================================================

\begin{table*}[ht]
\centering
\small
\setlength{\tabcolsep}{5pt}
\begin{tabular}{llccccccc}
\toprule
\textbf{DecoderLLM} & \textbf{Model} & \textbf{ContextEncoder} &
\textbf{$r$} & \textbf{EM} & \textbf{F1} \\
\midrule
\multirow{4}{*}{\shortstack{Llama-3\\8B-Instruct}}
 & Llama (baseline) & ---       & ---  & 0.6240 & 0.8097 \\
\cmidrule(lr){2-6}
 &  & OPT-2.7B  & 2/4/8 & 0.6570/{0.6680}/0.6670 & 0.8113/{0.8283}/0.8274 \\

 & CMC & OPT-1.3B  & 2/4/8 & 0.6570/{0.6680}/0.6670 & 0.8113/{0.8283}/0.8274 \\

 &  & GPT2-Large & 2/4/8 & 0.6530/\textbf{0.6700}/0.6670 & 0.8107/\textbf{0.8286}/0.8274 \\

\midrule
\multirow{4}{*}{\shortstack{Mistral-7B\\Instruct-v0.3}}
 & Mistral (baseline) & ---       & ---  & 0.6100 & 0.7586 \\
\cmidrule(lr){2-6}
 &  & OPT-2.7B  & 2/4/8 & 0.5740/0.5990/0.6000  & 0.7391/0.7533/0.7532 \\

 & CMC & OPT-1.3B  & 2/4/8 & 0.5740/0.5990/0.6000  & 0.7391/0.7533/0.7532 \\

 &  & GPT2-Large & 2/4/8 & 0.5740/\textbf{0.6020}/0.6000 & 0.7390/\textbf{0.7551}/0.7532 \\

\midrule
\multirow{4}{*}{\shortstack{Gemma-2\\9B-IT}}
 & Gemma (baseline) & ---       & ---  & 0.5510 & 0.7805 \\
\cmidrule(lr){2-6}
 &  & OPT-2.7B  & 2/4/8 & 0.6090/0.6100/\textbf{0.6110}  & 0.8102/0.8115/\textbf{0.8118}  \\

 & CMC & OPT-1.3B  & 2/4/8 & 0.6090/0.6100/0.6100 & 0.8102/0.8115/0.8115 \\

 &  & GPT2-Large & 2/4/8 & 0.6090/0.6100/0.6100 & 0.8102/0.8115/0.8115 \\

\bottomrule
\end{tabular}
\caption{EM and F1 results on \textbf{SQuAD} (controlled-data setting).}
\label{tab:squad_results}
\end{table*}

%% ============================================================
%% TABLE 2: AdversarialQA
%% ============================================================

\begin{table*}[ht]
\centering
\small
\setlength{\tabcolsep}{4pt}

\begin{tabular}{llccccc}
\toprule
\textbf{DecoderLLM} & \textbf{Model} & \textbf{ContextEncoder} &
\textbf{$r$} & \textbf{EM} & \textbf{F1} \\
\midrule
\multirow{4}{*}{\shortstack{Llama-3\\8B-Instruct}}
 & Llama (baseline) & ---        & ---  & 0.3430 & 0.5329 \\
\cmidrule(lr){2-6}
 &  & OPT-2.7B  & 2/4/8 & {0.3740}/0.3660/0.3670  & 0.5328/{0.5501}/0.5495 \\

 & CMC & OPT-1.3B  & 2/4/8 & {0.3740}/0.3660/0.3670  & 0.5328/ {0.5501}/0.5495 \\

 &  & GPT2-Large & 2/4/8 & \textbf{0.3750}/0.3680/0.3650  & 0.5330/\textbf{0.5509}/0.5464 \\

\midrule
\multirow{4}{*}{\shortstack{Mistral-7B\\Instruct-v0.3}}
 & Mistral (baseline) & ---        & ---  & 0.2960 & 0.4786 \\
\cmidrule(lr){2-6}
 &  & OPT-2.7B  & 2/4/8 & 0.3050/{0.3200}/{0.3200}  & 0.4888/{0.5013}/0.5012  \\

 & CMC & OPT-1.3B  & 2/4/8 & 0.3050/{0.3200}/{0.3200} & 0.4888/{0.5013}/0.5012 \\

 &  & GPT2-Large & 2/4/8 & 0.3060/\textbf{0.3220}/0.3180 & 0.4913/\textbf{0.5029}/0.5008 \\

\midrule
\multirow{4}{*}{\shortstack{Gemma-2\\9B-IT}}
 & Gemma (baseline) & ---        & ---  & 0.3290 & 0.5641 \\
\cmidrule(lr){2-6}
 &  & OPT-2.7B  & 2/4/8 & 0.3660/{0.3680}/{0.3680} & 0.5921/{0.5952}/{0.5952} \\

 & CMC & OPT-1.3B  & 2/4/8 & 0.3660/{0.3680}/{0.3680} & 0.5921/{0.5952}/{0.5952} \\

 &  & GPT2-Large & 2/4/8 & 0.3650/\textbf{0.3680}/{0.3680} & 0.5916/\textbf{0.5952}/{0.5952} \\

\bottomrule
\end{tabular}
\caption{EM and F1 results on \textbf{AdversarialQA} (controlled-data setting). %Full-context baseline vs.\ CMC with three ContextEncoders and compression rates $r \in \{2, 4, 8\}$.
}
\label{tab:adversarialqa_results}
\end{table*}

%% ============================================================
%% TABLE 3: HotpotQA
%% ============================================================
\begin{table*}[ht]
\centering
\small
\setlength{\tabcolsep}{4pt}
\begin{tabular}{llccccc}
\toprule
\textbf{DecoderLLM} & \textbf{Model} & \textbf{ContextEncoder} &
\textbf{$r$} & \textbf{EM} & \textbf{F1} \\
\midrule
\multirow{4}{*}{\shortstack{Llama-3\\8B-Instruct}}
 & Llama (baseline) & --- & --- & 0.4410 & 0.6814 \\
\cmidrule(lr){2-6}
 &  & OPT-2.7B  & 2/4/8 & 0.4610/0.4650/0.4610 & 0.6554/0.6634/0.6619 \\
 & CMC & OPT-1.3B  & 2/4/8 & 0.4880/0.4810/0.4790 & 0.6907/0.6933/0.6922 \\
 &  & GPT2-Large & 2/4/8 & \textbf{0.4960}/0.4840/0.4610 & 0.6934/\textbf{0.6943}/0.6619 \\
\midrule
\multirow{4}{*}{\shortstack{Mistral-7B\\Instruct-v0.3}}
 & Mistral (baseline) & --- & --- & 0.5300 & 0.7106 \\
\cmidrule(lr){2-6}
 &  & OPT-2.7B  & 2/4/8 & 0.4810/0.4910/0.4930 & 0.6540/0.6668/0.6672 \\
 & CMC & OPT-1.3B  & 2/4/8 & 0.5260/0.4910/\textbf{0.5380} & 0.7042/0.6668/\textbf{0.7169} \\
 &  & GPT2-Large & 2/4/8 & 0.4810/0.5360/\textbf{0.5380} & 0.6544/0.7158/\textbf{0.7169} \\
\midrule
\multirow{4}{*}{\shortstack{Gemma-2\\9B-IT}}
 & Gemma (baseline) & --- & --- & 0.4010 & 0.6700 \\
\cmidrule(lr){2-6}
 &  & OPT-2.7B  & 2/4/8 & {0.4200}/{0.4200}/0.4150 & {0.7059}/{0.7059}/0.7020 \\
 & CMC & OPT-1.3B  & 2/4/8 & {0.4200}/\textbf{0.4200}/0.4150 & {0.7059}/\textbf{0.7059}/0.7020 \\
 &  & GPT2-Large & 2/4/8 & {0.4200}/{0.4200}/0.4150 & {0.7059}/0.6949/0.6939 \\
\bottomrule
\end{tabular}
\caption{EM and F1 results on \textbf{HotpotQA} (controlled-data setting).
%Full-context baseline vs.\ CMC with three ContextEncoders and compression rates $r \in \{2, 4, 8\}$.
}
\label{tab:hotpotqa_results}
\end{table*}

%% ============================================================
%% TABLE 4: CovidQA
%% ============================================================
\begin{table*}[ht]
\centering
\small
\setlength{\tabcolsep}{4pt}

\begin{tabular}{llccccc}
\toprule
\textbf{DecoderLLM} & \textbf{Model} & \textbf{ContextEncoder} &
\textbf{$r$} & \textbf{EM} & \textbf{F1} \\
\midrule
\multirow{4}{*}{\shortstack{Llama-3\\8B-Instruct}}
 & Llama (baseline) & --- & --- & 0.2250 & 0.6383 \\
\cmidrule(lr){2-6}
 &  & OPT-2.7B  & 2/4/8 & {0.2950}/\textbf{0.2950}/{0.2950} & {0.6416}/\textbf{0.6416}/{0.6416} \\
 & CMC & OPT-1.3B  & 2/4/8 & 0.2900/0.2900/0.2900 & 0.6412/0.6392/0.6392 \\
 &  & GPT2-Large & 2/4/8 & 0.2900/0.2900/0.2900 & 0.6412/0.6392/0.6412 \\
\midrule
\multirow{4}{*}{\shortstack{Mistral-7B\\Instruct-v0.3}}
 & Mistral (baseline) & --- & --- & 0.1900 & 0.5761 \\
\cmidrule(lr){2-6}
 &  & OPT-2.7B  & 2/4/8 & 0.2150/{0.2200}/0.2150 & 0.6238/{0.6305}/0.6238 \\
 & CMC & OPT-1.3B  & 2/4/8 & 0.2150/{0.2200}/0.2150 & 0.6238/{0.6305}/0.6238 \\
 &  & GPT2-Large & 2/4/8 & 0.2150/\textbf{0.2200}/0.2150 & 0.6238/\textbf{0.6305}/0.6238 \\
 \midrule
\multirow{4}{*}{\shortstack{Gemma-2\\9B-IT}}
 & Gemma (baseline) & --- & --- & 0.1050 & 0.6209 \\
\cmidrule(lr){2-6}
 &  & OPT-2.7B  & 2/4/8 & 0.14/0.14/0.14 & 0.6323/0.6323/0.6323   \\
 & CMC & OPT-1.3B & 2/4/8 & 0.14/0.14/0.14 & 0.6323/0.6323/0.6323  \\
 &  & GPT2-Large & 2/4/8 & 0.14/0.14/0.14 & 0.6323/0.6323/0.6323  \\
\bottomrule
\end{tabular}
\caption{EM and F1 results on \textbf{CovidQA}.
%Full-context baseline vs.\ CMC with three ContextEncoders and compression rates $r \in \{2, 4, 8\}$.
}
\label{tab:covidqa_results}
\end{table*}

\begin{table*}[ht]
\centering
\small
\setlength{\tabcolsep}{4pt}

\small
\setlength{\tabcolsep}{5pt}
\begin{tabular}{llccccccc}
\hline
& &
\multicolumn{2}{c}{$r = 2$} &
\multicolumn{2}{c}{$r = 4$} &
\multicolumn{2}{c}{$r = 8$} \\
\cmidrule(lr){3-4}\cmidrule(lr){5-6}\cmidrule(lr){7-8}
\textbf{DecoderLLM} & \textbf{ContextEncoder} &
EM & F1 & EM & F1 & EM & F1 \\
\midrule
\multirow{3}{*}{\shortstack[l]{Llama-3\\8B-Instruct}}
 & OPT-2.7B   & 0.4468 & 0.6603 & 0.4485 & 0.6709 & 0.4475 & 0.6701 \\
 & OPT-1.3B   & 0.4522 & 0.6690 & 0.4512 & 0.6777 & 0.4507 & 0.6771 \\
 & GPT2-Large & {0.4535} & 0.6696 & 0.4530 & {0.6783} & 0.4457 & 0.6692 \\
\midrule
\multirow{3}{*}{\shortstack[l]{Mistral-7B\\Instruct-v0.3}}
 & OPT-2.7B   & 0.3937 & 0.6264 & 0.4075 & 0.6380 & 0.4070 & 0.6363 \\
 & OPT-1.3B   & 0.4050 & 0.6390 & 0.4075 & 0.6380 & 0.4183 & 0.6488 \\
 & GPT2-Large & 0.3940 & 0.6271 & {0.4200} & {0.6511} & 0.4178 & 0.6487 \\
\midrule
\multirow{3}{*}{\shortstack[l]{Gemma-2\\9B-IT}}
 & OPT-2.7B   & 0.3837 & 0.6851 & {0.3845} & {0.6862} & 0.3835 & 0.6853 \\
 & OPT-1.3B   & 0.3837 & 0.6851 & {0.3845} & {0.6862} & 0.3832 & 0.6852 \\
 & GPT2-Large & 0.3835 & 0.6850 & {0.3845} & 0.6835 & 0.3832 & 0.6832 \\

\midrule
\multicolumn{2}{l}{\textbf{Overall average}}
 & 0.4107 & 0.6607
 & \textbf{0.4157} & \textbf{0.6678}
 & 0.4152 & 0.6671 \\
\hline
\end{tabular}
\caption{Average EM and F1 across four datasets (SQuAD,
AdversarialQA, HotpotQA, CovidQA) under controlled-data setting, grouped by DecoderLLM, ContextEncoder, and compression rate $r \in \{2, 4, 8\}$.
\textbf{Bold} denotes the best average per DecoderLLM--ContextEncoder pair.}
\label{tab:avg_results}
\end{table*}

\subsection{Detailed Results per Dataset}
\label{app:detailed_results}

Tables~\ref{tab:squad_results}--\ref{tab:covidqa_results}
report EM and F1 for all nine encoder-decoder
combinations across $r \in \{2, 4, 8\}$ under
the controlled-data setting. \textbf{Bold} denotes
the best CMC result per DecoderLLM group. Three findings across these tables support the
design choices made in CMC. First, OPT-1.3B and
OPT-2.7B produce near-identical results across
all datasets and decoders, while GPT2-Large ---
the smallest encoder --- matches or outperforms
both, confirming that ContextEncoder scale does
not drive performance. Second, Gemma shows
near-zero variation across all encoder and
compression rate combinations ($\leq$0.001 EM),
indicating that the two-tier KV cache policy is
effective regardless of decoder sensitivity to
compression granularity. Third, CovidQA shows
consistent improvements across all nine
configurations despite only 1{,}500 training
samples, demonstrating that the two-phase
training strategy generalises to low-resource
domain-specific settings.

Additionally, Table~\ref{tab:avg_results} reports average EM and F1 across all encoder-decoder combinations and datasets. The overall average confirms $r=4$ as the best compression rate (EM~$=$~0.4157, F1~$=$~0.6678), marginally outperforming $r=8$ (EM~$=$~0.4152) and more substantially outperforming $r=2$ (EM~$=$~0.4107).

%% ============================================================
%% TABLE A: Llama-3-8B-Instruct — Inference Time
%% ============================================================

\begin{table*}[ht]
\centering
\small
\setlength{\tabcolsep}{5pt}
\begin{tabular}{clcrrrr}
\toprule
\textbf{Samples} & \textbf{Model} & \textbf{$T$} &
\textbf{Com.~(s)} & \textbf{Pre.~(s)} &
\textbf{Dec.~(s)} & \textbf{Total~(s)} \\
\midrule
\multirow{8}{*}{256}
 & Baseline & \multirow{2}{*}{512}  & -  &  7.61 &  1{,}535.77 &  1{,}543.38 \\
 & CMC      &   & 6.29  &  6.79 &  1{,}464.63 &  1{,}477.71 \\
 & Baseline & \multirow{2}{*}{1024}  & -  &  7.71 &  3{,}194.18 &  3{,}201.91 \\
 & CMC      &  & 6.21  &  6.87 &  2{,}930.89 &  2{,}943.98 \\
 & Baseline & \multirow{2}{*}{2048}  & - &  7.40 &  5{,}815.22 &  5{,}822.62 \\
 & CMC      &  & 6.60  &  6.31 &  5{,}067.51 &  5{,}080.43 \\
 & Baseline & \multirow{2}{*}{3000}  & -  &  7.61 & 10{,}717.09 & 10{,}724.70 \\
 & CMC      &  & 6.50  &  6.89 &  8{,}576.12 &  8{,}589.51 \\
\midrule
\multirow{8}{*}{512}
 & Baseline & \multirow{2}{*}{512}   & - & 16.60 &  2{,}574.88 &  2{,}591.49 \\
 & CMC      &   & 12.78 & 16.59 &  2{,}445.59 &  2{,}474.97 \\
 & Baseline & \multirow{2}{*}{1024}  & -  & 17.51 &  5{,}353.99 &  5{,}371.51 \\
 & CMC      &  & 12.96 & 12.61 &  4{,}931.13 &  4{,}956.70 \\
 & Baseline & \multirow{2}{*}{2048}  & -  & 14.88 & 11{,}639.93 & 11{,}654.82 \\
 & CMC      &  & 13.49 & 12.74 & 10{,}496.74 & 10{,}522.99 \\
 & Baseline & \multirow{2}{*}{3000}  & - & 17.67 & 18{,}233.12 & 18{,}250.79 \\
 & CMC      &  & 12.82 & 15.07 & 14{,}312.29 & 14{,}340.18 \\
\midrule
\multirow{8}{*}{1000}
 & Baseline & \multirow{2}{*}{512}   & - & 31.57 &  6{,}011.68 &  6{,}043.28 \\
 & CMC      &   & 25.14 & 26.94 &  5{,}711.32 &  5{,}763.45 \\
 & Baseline & \multirow{2}{*}{1024}  & -  & 31.46 & 12{,}498.05 & 12{,}529.52 \\
 & CMC      &  & 25.26 & 27.11 & 11{,}451.62 & 11{,}504.01 \\
 & Baseline & \multirow{2}{*}{2048}  & - & 31.62 & 26{,}954.16 & 26{,}985.80 \\
 & CMC      &  & 25.51 & 27.75 & 22{,}958.75 & 23{,}012.03 \\
 & Baseline & \multirow{2}{*}{3000}  & -  & 31.37 & 41{,}931.82 & 41{,}963.20 \\
 & CMC      &  & 25.69 & 27.13 & 33{,}508.74 & 33{,}561.58 \\
\bottomrule
\end{tabular}
\caption{Stage-wise inference time (seconds) for
\textbf{Llama-3-8B-Instruct} on SQuAD across three
sample sizes (GPT2-Large ContextEncoder, $r=4$, batch
size 8). Com.~=~compression; Pre.~=~prefill;
Dec.~=~decode. Baseline compression time is zero.}
\label{tab:llama_time}
\end{table*}

\subsection{Inference Efficiency}
\label{app:detailed_results_efficiency}

Tables~\ref{tab:llama_time}--\ref{tab:mistral_energy}
report stage-wise inference time and energy for
Llama and Mistral on SQuAD across three sample
sizes and $T \in \{512, 1024, 2048, 3000\}$.
%(GPT2-Large, $r=4$, batch size 8). 
These tables extend the 1{,}000-sample Llama results in Section~\ref{CMC_efficiency} to smaller evaluation
sets and Mistral. We present two additional observations. First, the compression overhead scales linearly with sample size --- approximately 6\,s at 256 samples, 13\,s at 512, and 26\,s at 1{,}000 --- and is independent of $T$, confirming that CMC's
compression cost is determined by the number of
contexts processed rather than the generation
length. Second, CMC prefill time is marginally
lower than the baseline across all conditions,
reflecting the shorter local context window relative to the uncompressed prompt. Moreover, Mistral shows consistent savings matching Llama ($\approx$20\% at $T=3{,}000$, 1{,}000 samples), confirming that the efficiency gains are decoder-independent.

%% ============================================================
%% Appendix: Inference Time and Energy Tables
%% Dataset: SQuAD | Encoder: GPT2-large | Batch: 8 | r=4
%% Sample sizes: 256, 512, 1000
%% ============================================================

%% ============================================================
%% TABLE B: Llama-3-8B-Instruct — Energy Consumption
%% ============================================================

\begin{table*}[ht]
\centering
\small
\setlength{\tabcolsep}{5pt}
\begin{tabular}{clccccc}
\toprule
\textbf{Samples} & \textbf{Model} & \textbf{$T$} &
\textbf{Com.~(kWh)} & \textbf{Pre.~(kWh)} &
\textbf{Dec.~(kWh)} & \textbf{Total~(kWh)} \\
\midrule
\multirow{8}{*}{256}
 & Baseline & \multirow{2}{*}{512}   & - & 0.0012 & 0.2392 & 0.2404 \\
 & CMC      &   & 0.0010 & 0.0011 & 0.2278 & 0.2298 \\
 & Baseline & \multirow{2}{*}{1024}  & - & 0.0012 & 0.4966 & 0.4978 \\
 & CMC      &  & 0.0010 & 0.0011 & 0.4555 & 0.4576 \\
 & Baseline & \multirow{2}{*}{2048}  & - & 0.0017 & 1.3063 & 1.3079 \\
 & CMC      &  & 0.0014 & 0.0013 & 1.0681 & 1.0709 \\
 & Baseline & \multirow{2}{*}{3000}  & - & 0.0012 & 1.7187 & 1.7200 \\
 & CMC      &  & 0.0010 & 0.0011 & 1.3696 & 1.3718 \\
\midrule
\multirow{8}{*}{512}
 & Baseline & \multirow{2}{*}{512}   & - & 0.0038 & 0.5900 & 0.5938 \\
 & CMC      &   & 0.0029 & 0.0037 & 0.5496 & 0.5562 \\
 & Baseline & \multirow{2}{*}{1024}  & - & 0.0040 & 1.2227 & 1.2267 \\
 & CMC      &  & 0.0029 & 0.0028 & 1.1025 & 1.1082 \\
 & Baseline & \multirow{2}{*}{2048}  & - & 0.0033 & 2.6022 & 2.6056 \\
 & CMC      &  & 0.0028 & 0.0027 & 2.2001 & 2.2056 \\
 & Baseline & \multirow{2}{*}{3000}  & - & 0.0040 & 4.0996 & 4.1036 \\
 & CMC      &  & 0.0028 & 0.0033 & 3.1068 & 3.1128 \\
\midrule
\multirow{8}{*}{1000}
 & Baseline & \multirow{2}{*}{512}   & - & 0.0050 & 0.9514 & 0.9564 \\
 & CMC      &   & 0.0040 & 0.0043 & 0.9043 & 0.9125 \\
 & Baseline & \multirow{2}{*}{1024}  & - & 0.0050 & 1.9723 & 1.9773 \\
 & CMC      &  & 0.0040 & 0.0043 & 1.8009 & 1.8092 \\
 & Baseline & \multirow{2}{*}{2048}  & - & 0.0050 & 4.2575 & 4.2625 \\
 & CMC      &  & 0.0040 & 0.0044 & 3.6124 & 3.6208 \\
 & Baseline & \multirow{2}{*}{3000}  & - & 0.0050 & 6.7204 & 6.7254 \\
 & CMC      &  & 0.0041 & 0.0043 & 5.3538 & 5.3623 \\
\bottomrule
\end{tabular}
\caption{Stage-wise energy consumption (kWh) for \textbf{Llama-3-8B-Instruct} on SQuAD across three sample sizes (GPT2-Large ContextEncoder, $r=4$, batch size 8). Stage-wise energy is apportioned proportionally to wall-clock time.}
\label{tab:llama_energy}
\end{table*}

%% ============================================================
%% TABLE C: Mistral-7B-Instruct-v0.3 — Inference Time
%% ============================================================

\begin{table*}[ht]
\centering
\small
\setlength{\tabcolsep}{5pt}
\begin{tabular}{clcrrrr}
\toprule
\textbf{Samples} & \textbf{Model} & \textbf{$T$} &
\textbf{Com.~(s)} & \textbf{Pre.~(s)} &
\textbf{Dec.~(s)} & \textbf{Total~(s)} \\
\midrule
\multirow{8}{*}{256}
 & Baseline & \multirow{2}{*}{512}   & -  &  8.64 &  1{,}537.28 &  1{,}545.92 \\
 & CMC      &   & 6.42  &  7.27 &  1{,}461.70 &  1{,}475.39 \\
 & Baseline & \multirow{2}{*}{1024}  &  - &  8.44 &  3{,}190.70 &  3{,}199.14 \\
 & CMC      &  & 6.33  &  7.03 &  2{,}919.23 &  2{,}932.59 \\
 & Baseline & \multirow{2}{*}{2048}  & - &  8.47 &  6{,}864.73 &  6{,}873.19 \\
 & CMC      &  & 6.41  &  6.94 &  5{,}846.33 &  5{,}859.69 \\
 & Baseline & \multirow{2}{*}{3000}  & -  &  8.86 &  9{,}094.45 &  9{,}103.32 \\
 & CMC      &  & 6.42  &  6.21 &  7{,}429.81 &  7{,}442.45 \\
\midrule
\multirow{8}{*}{512}
 & Baseline & \multirow{2}{*}{512}   & -  & 16.73 &  3{,}061.61 &  3{,}078.34 \\
 & CMC      &   & 12.08 & 14.14 &  2{,}911.43 &  2{,}937.66 \\
 & Baseline & \multirow{2}{*}{1024}  & -  & 16.75 &  5{,}415.73 &  5{,}432.49 \\
 & CMC      &  & 12.94 & 12.42 &  5{,}179.11 &  5{,}204.47 \\
 & Baseline & \multirow{2}{*}{2048}  &  - & 15.98 & 11{,}465.26 & 11{,}481.24 \\
 & CMC      &  & 13.03 & 12.34 & 10{,}250.06 & 10{,}275.45 \\
 & Baseline & \multirow{2}{*}{3000}  &  - & 15.54 & 18{,}180.74 & 18{,}196.29 \\
 & CMC      &  & 12.91 & 13.25 & 14{,}883.97 & 14{,}910.14 \\
\midrule
\multirow{8}{*}{1000}
 & Baseline & \multirow{2}{*}{512}   & -  & 33.08 &  5{,}179.30 &  5{,}212.39 \\
 & CMC      &   & 26.00 & 28.63 &  5{,}058.98 &  5{,}113.63 \\
 & Baseline & \multirow{2}{*}{1024}  & -  & 31.34 & 10{,}657.52 & 10{,}688.87 \\
 & CMC      &  & 27.17 & 24.45 & 10{,}295.92 & 10{,}347.56 \\
 & Baseline & \multirow{2}{*}{2048}  & -  & 34.24 & 26{,}854.94 & 26{,}889.19 \\
 & CMC      &  & 24.87 & 28.05 & 22{,}832.35 & 22{,}885.29 \\
 & Baseline & \multirow{2}{*}{3000}  & -  & 34.29 & 41{,}807.04 & 41{,}841.34 \\
 & CMC      &  & 26.09 & 27.79 & 33{,}354.42 & 33{,}408.33 \\
\bottomrule
\end{tabular}
\caption{Stage-wise inference time (seconds) for
\textbf{Mistral-7B-Instruct-v0.3} on SQuAD across
three sample sizes (GPT2-Large ContextEncoder,
$r=4$, batch size 8). Com.~=~compression;
Pre.~=~prefill; Dec.~=~decode.}
\label{tab:mistral_time}
\end{table*}

%% ============================================================
%% TABLE D: Mistral-7B-Instruct-v0.3 — Energy Consumption
%% ============================================================

\begin{table*}[ht]
\centering
\small
\setlength{\tabcolsep}{5pt}
\begin{tabular}{clccccc}
\toprule
\textbf{Samples} & \textbf{Model} & \textbf{$T$} &
\textbf{Com.~(kWh)} & \textbf{Pre.~(kWh)} &
\textbf{Dec.~(kWh)} & \textbf{Total~(kWh)} \\
\midrule
\multirow{8}{*}{256}
 & Baseline & \multirow{2}{*}{512}   & - & 0.0014 & 0.2464 & 0.2478 \\
 & CMC      &   & 0.0010 & 0.0012 & 0.2335 & 0.2357 \\
 & Baseline & \multirow{2}{*}{1024}  & - & 0.0014 & 0.5199 & 0.5213 \\
 & CMC      &  & 0.0010 & 0.0011 & 0.4727 & 0.4749 \\
 & Baseline & \multirow{2}{*}{2048}  & - & 0.0014 & 1.1076 & 1.1089 \\
 & CMC      &  & 0.0010 & 0.0011 & 0.9351 & 0.9373 \\
 & Baseline & \multirow{2}{*}{3000}  & - & 0.0020 & 2.0874 & 2.0895 \\
 & CMC      &  & 0.0013 & 0.0013 & 1.6039& 1.6066 \\
\midrule
\multirow{8}{*}{512}
 & Baseline & \multirow{2}{*}{512}   & - & 0.0027 & 0.4966 & 0.4993 \\
 & CMC      &   & 0.0020 & 0.0023 & 0.4713 & 0.4756 \\
 & Baseline & \multirow{2}{*}{1024}  & - & 0.0037 & 1.1919 & 1.1956 \\
 & CMC      &  & 0.0028 & 0.0027 & 1.1356 & 1.1411 \\
 & Baseline & \multirow{2}{*}{2048}  & - & 0.0037 & 2.6172 & 2.6209 \\
 & CMC      &  & 0.0027 & 0.0026 & 2.1315 & 2.1367 \\
 & Baseline & \multirow{2}{*}{3000}  & - & 0.0036 & 4.1707 & 4.1743 \\
 & CMC      &  & 0.0028 & 0.0029 & 3.2180 & 3.2237 \\
\midrule
\multirow{8}{*}{1000}
 & Baseline & \multirow{2}{*}{512}   & - & 0.0072 & 1.1238 & 1.1310 \\
 & CMC      &   & 0.0054 & 0.0060 & 1.0550 & 1.0664 \\
 & Baseline & \multirow{2}{*}{1024}  & - & 0.0068 & 2.3237 & 2.3306 \\
 & CMC      &  & 0.0055 & 0.0050 & 2.1009 & 2.1115 \\
 & Baseline & \multirow{2}{*}{2048}  & - & 0.0055 & 4.3293 & 4.3349 \\
 & CMC      &  & 0.0040 & 0.0045 & 3.6503 & 3.6587 \\
 & Baseline & \multirow{2}{*}{3000}  & - & 0.0056 & 6.7946 & 6.8001 \\
 & CMC      &  & 0.0042 & 0.0044 & 5.3397 & 5.3483 \\
\bottomrule
\end{tabular}
\caption{Stage-wise energy consumption (kWh) for
\textbf{Mistral-7B-Instruct-v0.3} on SQuAD across
three sample sizes (GPT2-Large ContextEncoder,
$r=4$, batch size 8). Stage-wise energy is apportioned proportionally
to wall-clock time.}
\label{tab:mistral_energy}
\end{table*}

\subsection{Peak GPU Memory}
\label{app:memory_appendix}
Tables~\ref{tab:llama_memory} and~\ref{tab:mistral_memory} report peak allocated and peak reserved GPU memory for Llama and Mistral across three sample sizes and $T \in \{512, 1024, 2048, 3000\}$, extending the 1{,}000-sample Llama results in Section~\ref{CMC_efficiency}. Peak allocated memory is stable across $T$ for
CMC ($\approx$8.85\,GB for Llama, $\approx$8.43\,GB for Mistral) regardless of sample size. The baseline is also stable for Llama ($\approx$10.0\,GB) but grows with $T$ for Mistral (9.1\,GB at $T=512$ to 11.1\,GB at $T=3{,}000$), reflecting Mistral's longer tokenised prompts and larger KV cache growth per step. The key finding is in peak reserved memory. CMC maintains a near-constant footprint across all sample sizes and token budgets ($\approx$9.8\,GB for Llama, $\approx$9.1\,GB for Mistral), while the baseline spikes sharply at large $T$: up to 19.7\,GB for Llama and 24.3\,GB for Mistral at $T=3{,}000$. This confirms that the bounded two-tier KV cache prevents memory growth regardless of decoder architecture, sample size, or generation length.

\begin{table*}[ht]
\centering
\small
\setlength{\tabcolsep}{5pt}
\begin{tabular}{clccccc}
\toprule
& & &
\multicolumn{2}{c}{\textbf{Peak Alloc.~(GB)}} &
\multicolumn{2}{c}{\textbf{Peak Res.~(GB)}} \\
\cmidrule(lr){4-5}\cmidrule(lr){6-7}
\textbf{Samples} & \textbf{Model} & \textbf{$T$} &
\textbf{Base} & \textbf{CMC} &
\textbf{Base} & \textbf{CMC} \\
\midrule
\multirow{4}{*}{256}
 & \multirow{4}{*}{\shortstack{Baseline\\vs\\CMC}}
   & 512  & 10.033 & 8.916 & 11.234 &  9.742 \\
 & & 1024 &  9.968 & 8.848 & 11.125 &  9.742 \\
 & & 2048 &  9.970 & 8.806 & 11.100 &  9.717 \\
 & & 3000 & 10.151 & 8.848 & 15.332 &  9.742 \\
\midrule
\multirow{4}{*}{512}
 & \multirow{4}{*}{\shortstack{Baseline\\vs\\CMC}}
   & 512  & 10.018 & 8.850 & 11.172 &  9.807 \\
 & & 1024 & 10.027 & 8.859 & 12.436 & 11.381 \\
 & & 2048 & 10.018 & 8.850 & 11.172 &  9.807 \\
 & & 3000 & 10.160 & 8.850 & 18.617 &  9.807 \\
\midrule
\multirow{4}{*}{1000}
 & \multirow{4}{*}{\shortstack{Baseline\\vs\\CMC}}
   & 512  & 10.017 & 8.853 & 11.197 &  9.836 \\
 & & 1024 & 10.017 & 8.853 & 11.197 &  9.836 \\
 & & 2048 & 10.017 & 8.853 & 11.199 &  9.838 \\
 & & 3000 & 10.159 & 8.853 & 19.689 &  9.836 \\
\bottomrule
\end{tabular}
\caption{Peak allocated and peak reserved GPU
memory (GB) for \textbf{Llama-3-8B-Instruct} on
SQuAD (GPT2-Large, $r=4$, batch size 8). Peak
allocated memory is measured via
\texttt{torch.cuda.max\_memory\_allocated()} and
peak reserved memory via
\texttt{torch.cuda.max\_memory\_reserved()}.}
\label{tab:llama_memory}
\end{table*}

%% ============================================================
%% TABLE F: Mistral-7B-Instruct-v0.3 — Peak GPU Memory
%% ============================================================

\begin{table*}[ht]
\centering
\small
\setlength{\tabcolsep}{5pt}
\begin{tabular}{clccccc}
\toprule
& & &
\multicolumn{2}{c}{\textbf{Peak Alloc.~(GB)}} &
\multicolumn{2}{c}{\textbf{Peak Res.~(GB)}} \\
\cmidrule(lr){4-5}\cmidrule(lr){6-7}
\textbf{Samples} & \textbf{Model} & \textbf{$T$} &
\textbf{Base} & \textbf{CMC} &
\textbf{Base} & \textbf{CMC} \\
\midrule
\multirow{4}{*}{256}
 & \multirow{4}{*}{\shortstack{Baseline\\vs\\CMC}}
   & 512  &  9.052 & 8.432 & 10.197 & 9.084 \\
 & & 1024 &  9.266 & 8.434 & 10.346 & 9.084 \\
 & & 2048 & 10.203 & 8.434 & 14.162 & 9.084 \\
 & & 3000 & 11.075 & 8.397 & 24.262 & 9.082 \\
\midrule
\multirow{4}{*}{512}
 & \multirow{4}{*}{\shortstack{Baseline\\vs\\CMC}}
   & 512  &  9.073 & 8.449 & 10.434 & 9.146 \\
 & & 1024 &  9.276 & 8.431 & 10.328 & 9.082 \\
 & & 2048 & 10.192 & 8.431 & 14.125 & 9.082 \\
 & & 3000 & 11.085 & 8.431 & 24.262 & 9.082 \\
\midrule
\multirow{4}{*}{1000}
 & \multirow{4}{*}{\shortstack{Baseline\\vs\\CMC}}
   & 512  &  9.074 & 8.431 & 10.221 & 9.082 \\
 & & 1024 &  9.276 & 8.431 & 10.328 & 9.082 \\
 & & 2048 & 10.213 & 8.449 & 14.166 & 9.109 \\
 & & 3000 & 11.084 & 8.449 & 24.303 & 9.109 \\
\bottomrule
\end{tabular}
\caption{Peak allocated and peak reserved GPU
memory (GB) for \textbf{Mistral-7B-Instruct-v0.3}
on SQuAD (GPT2-Large, $r=4$, batch size 8). Peak
reserved memory is measured via
\texttt{torch.cuda.max\_memory\_reserved()}.}
\label{tab:mistral_memory}
\end{table*}

\subsection{Attention MACs Analysis}
\label{app:macs_appendix}
Attention MACs are estimated analytically as
$\text{MACs}_t = 2 \cdot n_h \cdot d_h \cdot n_l
\cdot L_t$ per generation step, %where $L_t \leq K + W$ for CMC vs.\ $L_t = P + t$ for the baseline. 
where $L_t \le K+W+|q|$ for CMC vs. $L_t = P+t$ for the baseline, with $P$ denoting the combined length of the baseline's context window and question, and $t$ the number of tokens generated so far. Absolute values differ across decoders due to architecture ($n_l=42$ for Gemma vs.\ 32 for Llama and Mistral) and tokeniser-dependent prompt lengths. Table~\ref{tab:macs} shows that the $K{+}W$ bound
reduces attention MACs by 27.0\% on average
(6.6\%--51.3\%) across all nine combinations.
Reductions are largest on SQuAD (34.0\% average)
where longer prompt lengths make the fixed cache
bound more effective. Smaller reductions on
HotpotQA for Llama and Gemma (6.6\% and 8.3\%)
reflect shorter baseline prompts in that split.
These reductions are consistent with the empirical
efficiency gains in Section~\ref{CMC_efficiency},
confirming that CMC's efficiency advantage is
architectural rather than hardware-specific.

\begin{table*}[ht]
\centering
\small
\setlength{\tabcolsep}{5pt}
\begin{tabular}{llccc}
\toprule
\textbf{Dataset} & \textbf{Decoder} &
\textbf{Baseline} & \textbf{CMC} & \textbf{Reduction} \\
\midrule
\multirow{3}{*}{SQuAD}
  & Llama   & 1.61 & 1.12 & 30.6\% \\
  & Mistral & 1.82 & 1.12 & 38.1\% \\
  & Gemma   & 3.00 & 2.01 & 33.0\% \\
\midrule
\multirow{3}{*}{AdversarialQA}
  & Llama   & 0.71 & 0.55 & 22.0\% \\
  & Mistral & 0.77 & 0.56 & 27.4\% \\
  & Gemma   & 2.58 & 1.94 & 24.9\% \\
\midrule
\multirow{3}{*}{HotpotQA}
  & Llama   & 0.60 & 0.56 &  6.6\% \\
  & Mistral & 0.72 & 0.35 & 51.3\% \\
  & Gemma   & 1.56 & 1.43 &  8.3\% \\
\bottomrule
\end{tabular}
\caption{Estimated attention MACs (Billion) for CMC
vs.\ the baseline ($r=4$,
$W=130$, $K=40$).}
\label{tab:macs}
\end{table*}

\begin{comment}

\subsection{Top-$K$ CME Selection Sensitivity}
\label{app:topk}

Figure~\ref{fig:topk_sensitivity} shows EM and F1 as a function of $K$ on SQuAD (2{,}000 validation
samples, GPT2-Large, $r=4$, Llama, $W=140$). EM decreases monotonically with $K$: from 0.555 at $K=10$ to 0.482 at $K=20$, then plateauing at $\approx$0.407 for $K \geq 30$. The plateau arises because $K$ exceeds the total CME count $M_b$ for most SQuAD passages at this compression setting, making top-$K$ selection equivalent to supplying all CMEs without filtering. The gradient across $K=10$, $K=20$, and $K=30$ confirms that each additional CME beyond the most informative ones introduces noise into the Tier-1 prefix, progressively diluting the answer-relevant signal. This is consistent with the ablation result that replacing query-guided selection with random CME selection (A2) reduces EM by 8.1 points. 

\end{comment}

\subsection{Local Window Size Sensitivity}
\label{app:window}

\begin{figure}[ht]
\centering
\begin{tikzpicture}
\begin{axis}[
    width=\linewidth,
    height=5.5cm,
    xlabel={Local window $W$ (tokens)},
    ylabel={Score},
    xmin=35, xmax=270,
    ymin=0.0, ymax=0.75,
    xtick={50,80,100,130,160,200,256},
    ytick={0.0,0.1,0.2,0.3,0.4,0.5,0.6,0.7},
    yticklabel={\pgfmathprintnumber[fixed,precision=1]{\tick}},
    grid=both,
    grid style={line width=0.3pt, draw=gray!20},
    major grid style={line width=0.4pt, draw=gray!30},
    tick align=outside,
    tick pos=left,
    legend style={
        at={(0.05,0.98)},
        anchor=north west,
        font=\small,
        draw=gray!40,
        fill=white,
        row sep=2pt,
    },
    label style={font=\small},
    tick label style={font=\small},
    clip=false,
]

%% ── EM line ──────────────────────────────────────────────
\addplot[
    color=orange,
    mark=*,
    mark size=2.5pt,
    line width=1.2pt,
    mark options={fill=orange, draw=orange},
] coordinates {
(50,0.14)(80,0.2335)(100,0.27)
    (130,0.3040)(160,0.318)(200,0.354)(256,0.3545)
};
\addlegendentry{EM}

%% ── F1 line ──────────────────────────────────────────────
\addplot[
    color=Green,
    mark=square*,
    mark size=2.0pt,
    line width=1.2pt,
    dashed,
    mark options={fill=Green, draw=Green},
] coordinates {
(50,0.2574)(80,0.4012)(100,0.4501)
    (130,0.4981)(160,0.5108)(200,0.5402)(256,0.5410)
};
\addlegendentry{F1}

\end{axis}
\end{tikzpicture}
\caption{Local window $W$ sensitivity on SQuAD
(full-data setting, GPT2-Large, $r=4$, Llama,
$K=30$, 2000 validation samples). EM and F1
increase monotonically with $W$, plateauing at
$W \geq 200$. %The default $W=130$ (dashed vertical line) is 6 EM points below the optimal $W=200$, indicating that larger windows benefit span extraction when memory budget permits.
}
\label{fig:window_sensitivity}
\end{figure}
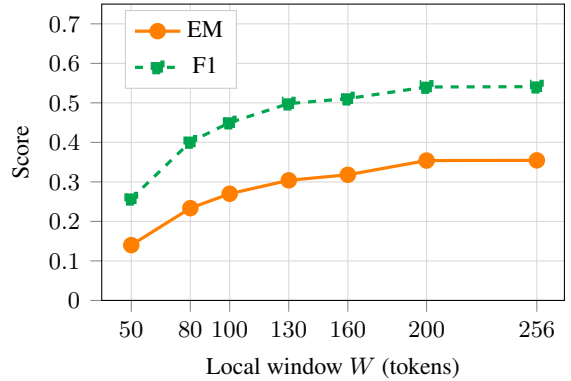

Figure~\ref{fig:window_sensitivity} shows EM and
F1 as a function of the Tier-2 local window size
$W$ on SQuAD. EM increases monotonically from 0.140
at $W=50$ to 0.354 at $W=200$, then plateaus at
$W=256$ (EM~$=$~0.355), confirming that larger
local windows consistently benefit extractive QA
by providing more surrounding context for precise
span extraction. The plateau at $W \geq 200$
suggests diminishing returns once the window
captures the full answer region for most passages.
The default $W=130$ achieves EM~$=$~0.304, which
is 5.0 points below the  $W=200$
(EM~$=$~0.354), indicating that $W=200$ is a
stronger default for deployment when the additional
KV budget ($K + W = 230$ tokens) is available.
The finding is consistent with the 200-sample
evaluation, which shows the same monotonic pattern
and plateau boundary, confirming robustness to
sample size.

\subsection{Denoising Threshold ($\tau$) Sensitivity}
\label{app:tau}

\begin{figure}[ht]
\centering
\begin{tikzpicture}
\begin{axis}[
    width=\linewidth,
    height=5.5cm,
    xlabel={Denoising threshold $\tau$},
    ylabel={Score},
    xmin=-0.01, xmax=0.21,
    ymin=0.0, ymax=0.85,
    xtick={0.00,0.02,0.05,0.08,0.10,0.15,0.20},
    xticklabels={0.00,0.02,0.05,0.08,0.10,0.15,0.20},
    ytick={0.0,0.1,0.2,0.3,0.4,0.5,0.6,0.7,0.8},
    yticklabel={\pgfmathprintnumber[fixed,precision=1]{\tick}},
    grid=both,
    grid style={line width=0.3pt, draw=gray!20},
    major grid style={line width=0.4pt, draw=gray!30},
    tick align=outside,
    tick pos=left,
    legend style={
        at={(0.98,0.40)},
        anchor=north east,
        font=\small,
        draw=gray!40,
        fill=white,
        row sep=2pt,
    },
    label style={font=\scriptsize},
    tick label style={font=\scriptsize},
    clip=false,
]

%% ── EM line ──────────────────────────────────────────────
\addplot[
    color=Red,
    mark=*,
    mark size=2.5pt,
    line width=1.2pt,
    mark options={fill=Red, draw=Red},
] coordinates {
    (0.00, 0.3070)
    (0.02, 0.5065)
    (0.05, 0.5065)
    (0.08, 0.5065)
    (0.10, 0.5065)
    (0.15, 0.5065)
    (0.20, 0.5065)
};
\addlegendentry{EM}

%% ── F1 line ──────────────────────────────────────────────
\addplot[
    color=Blue,
    mark=square*,
    mark size=2.0pt,
    line width=1.2pt,
    dashed,
    mark options={fill=Blue, draw=Blue},
] coordinates {
    (0.00, 0.5005)
    (0.02, 0.7385)
    (0.05, 0.7385)
    (0.08, 0.7385)
    (0.10, 0.7385)
    (0.15, 0.7385)
    (0.20, 0.7385)
};
\addlegendentry{F1}

%% ── Default tau=0.05 vertical marker ────────────────────
\addplot[
    color=green,
    line width=0.8pt,
    dashed,
    forget plot,
] coordinates {(0.05, 0.0) (0.05, 0.80)};
\begin{comment}
\node[
    font=\scriptsize,
    text=gray!70,
    anchor=south,
] at (axis cs:0.05, 0.81) {default};

%% ── Stable region annotation ─────────────────────────────
\draw[
    color=gray!50,
    line width=0.5pt,
    decorate,
    decoration={brace, amplitude=3pt, mirror},
] (axis cs:0.02,-0.08) -- (axis cs:0.20,-0.08)
    node[midway, below=5pt, font=\scriptsize, text=gray!70]
    {stable ($\tau \geq 0.02$)};

%% ── No-denoising annotation ─────────────────────────────
\node[
    font=\scriptsize,
    text=red!60!black,
    anchor=north west,
] at (axis cs:0.00, 0.2940) {\;no denoising};
\end{comment}
\end{axis}
\end{tikzpicture}
\caption{Denoising threshold $\tau$ sensitivity on SQuAD (full-data setting, GPT2-Large, $r=4$, Llama, $K=30$, $W=140$, 2{,}000 validation
samples). No denoising ($\tau=0$) reduces EM by 19.9 points. For
$\tau \geq 0.02$, EM and F1 are identical as the \texttt{keep\_min\_every}$=16$ floor becomes the binding constraint. The default $\tau=0.05$ (dashed line) lies within the
stable region.}
\label{fig:tau_sensitivity}
\end{figure}
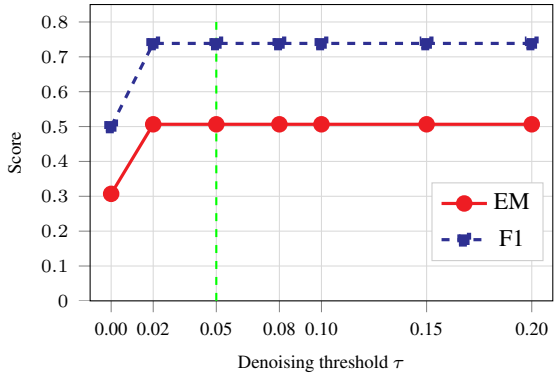

Figure~\ref{fig:tau_sensitivity} shows EM and F1
as a function of the graph-based context denoising
threshold $\tau$ on SQuAD. The results reveal a binary pattern. At $\tau=0$ (no denoising), EM drops to 0.307 --- a 19.9-point degradation relative to any denoising setting ($\tau \geq 0.02$, EM~$=$~0.507). This confirms that graph-based context denoising is critical for CME quality: without it, the ContextEncoder compresses all tokens including highly redundant ones, producing a noisier and less informative CME prefix. For $\tau \geq 0.02$, EM and F1 are identical across all tested values (EM~$=$~0.507, F1~$=$~0.739), with token retention stabilising at 14.0\%. This plateau arises because the \texttt{keep\_min\_every}~$=$~16 constraint -- which forces retention of at least one token every 16 positions -- becomes the binding constraint for $\tau \geq 0.02$. Once this floor dominates, the exact
value of $\tau$ has no further effect on which
tokens are retained. The default $\tau=0.05$ is therefore well within the stable region and robust to any choice in the range $[0.02, 0.20]$.

%% ── TikZ figure ───────────────────────────────────────────

\subsection{Tier-2 Window Localization Analysis}
\label{app:oracle_free}
In main experiments, the Tier-2 local window is
centered using the gold answer character offset
$\alpha$ provided by benchmark annotations, applied
identically to both CMC and the baseline so that $\alpha$ is a controlled
variable rather than information available only to
CMC. While this is standard practice in extractive QA
evaluation, $\alpha$ is unavailable at deployment
time. %Table~\ref{tab:oracle_free} quantifies the performance gap between the oracle offset and three deployment-realistic alternatives on 2{,}000 validation samples from SQuAD, AdversarialQA, and HotpotQA (GPT2-Large, $r=4$, Llama, $K=30$, $W=140$).
Table~\ref{tab:oracle_free} quantifies the
performance gap between the oracle offset and
three deployment-realistic alternatives on 2{,}000
validation samples from SQuAD and AdversarialQA
(full-data setting) and HotpotQA (controlled-data
setting) (GPT2-Large, $r=4$, Llama, $K=30$,
$W=140$). \noindent\textbf{BM25 top-1 sentence}
\cite{robertson2009probabilistic} ranks context
sentences by relevance to the question and uses the
midpoint of the top-ranked sentence as $\alpha$.
\noindent\textbf{CME-guided localization} uses the
character offset of the chunk that produced the
highest-norm CME as $\alpha$, requiring no external
library. \noindent\textbf{Midpoint heuristic} always
centres at $\alpha = |C|/2$, using no information
about the question or answer at all. 

On SQuAD, oracle-free localization stays close to
the oracle: BM25 costs only 0.4 EM points
($\Delta$EM~$=$~$-$0.004, 99\% retention),
CME-guided costs 2.0 points ($-$0.020, 95\%), and
even the midpoint heuristic retains 91\% of oracle
EM. HotpotQA shows the same pattern, and more
strongly: BM25, CME-guided, and the midpoint
heuristic all retain 97--98\% of oracle EM, with
CME-guided (0.415) and the midpoint heuristic
(0.414) essentially matching BM25 (0.412). We
attribute this robustness to HotpotQA's multi-hop
structure, where answer-supporting evidence is
distributed across multiple passages rather than
concentrated at a single span. %, so the precise window center matters less than having a localwindow at all. 
AdversarialQA shows the largest oracle dependence ($-$0.052 to $-$0.069, 78--84\% retention), consistent with its adversarially constructed questions being designed to resist the shallow lexical and positional cues that BM25 and the midpoint heuristic rely on. These experimental results show that CMC's Tier-2 benefit comes from the presence of a local context window, not from privileged access to the gold answer location: oracle-free localization retains the large majority of oracle performance (78$-$99\%) across single-hop, adversarial, and multi-hop QA datasets.

\begin{table*}[ht]
\centering
\small
\begin{tabular}{clccc}
\toprule
\textbf{Dataset}& \textbf{Tier-2 localisation} & \textbf{EM} &
\textbf{F1} & $\boldsymbol{\Delta}$\textbf{EM} \\
\midrule
\multirow{4}{*}{SQuAD} & \textbf{Oracle} ($\alpha$ from annotations)
    & \textbf{0.408} & \textbf{0.591} & --- \\
& BM25 top-1 sentence
    & 0.404 & 0.585 & $-$0.004 \\
& CME-guided (top-1 CME chunk)
    & 0.388 & 0.556 & $-$0.020 \\
& Midpoint heuristic
    & 0.373 & 0.533 & $-$0.035 \\ \hline
\multirow{4}{*}{AdversarialQA} & \textbf{Oracle}
    & \textbf{0.321} & \textbf{0.501} & --- \\
& BM25
    & 0.269 & 0.428 & $-$0.052 \\
& CME-guided 
    & 0.264 & 0.405 & $-$0.057 \\
& Midpoint heuristic
    & 0.252 & 0.389 & $-$0.069 \\ \hline
\multirow{4}{*}{HotpotQA} & \textbf{Oracle}
    & \textbf{0.423} & \textbf{0.634} & --- \\
& BM25
    & 0.412 & 0.621 & $-$0.011 \\
& CME-guided 
    & 0.415 & 0.622 & $-$0.008 \\
& Midpoint heuristic
    & 0.414 & 0.624 & $-$0.009 \\ \hline
\bottomrule
\end{tabular}
\caption{Effect of Tier-2 window localization
method on SQuAD (full-data), AdversarialQA (full-data), and HotpotQA (controlled-data) (GPT2-Large, $r=4$, Llama,
$K=30$, $W=140$, 2{,}000 validation samples per
dataset). $\Delta$EM is relative to the oracle
within each dataset.}
\label{tab:oracle_free}
\end{table*}

\subsection{Case Study}
\label{app:case_study}

Table~\ref{tab:case_study} presents four
representative examples from the SQuAD validation
set illustrating the behaviour of CMC relative to
the baseline across four outcome categories. \textbf{Case~1} shows a straightforward factual question
where both CMC and the baseline correctly extract the gold answer
(\textit{oxides}), confirming that CMC preserves
answer quality on simple single-token spans. \textbf{Case~2} demonstrates a clear CMC advantage. The context contains two monetary figures --- £43 million (the original gallery budget) and £76 million (the revised cost estimate). The baseline anchors to the more prominent earlier mention and returns the incorrect figure. The CME prefix supplies sufficient global context for CMC to distinguish between the two figures and return the correct answer. \textbf{Case~3}  illustrates a limitation of CMC. Both CMC and the baseline locate the correct passage, but CMC produces \textit{1979} rather than the gold \textit{June 1979}, losing the month through compression. This suggests that fine-grained temporal spans are susceptible to information loss when the relevant token is not well-represented in the CME prefix.  \textbf{Case~4}  shows a partial improvement. The gold answer is a full sentence (\textit{The city's residents fled to the north}), which neither model matches exactly. The baseline retrieves only \textit{the north} (F1~$=$~0.40), while CMC retrieves \textit{to the north} (F1~$=$~0.55), a longer span that better overlaps with the gold answer. This illustrates that CMC can improve partial answer coverage even when an exact match is not achieved.

\begin{table*}[ht]
\centering
\small
\setlength{\tabcolsep}{5pt}
\renewcommand{\arraystretch}{1.3}

\begin{tabular}{p{14cm}}

%% ── Case 1: Both correct ─────────────────────────────────
\toprule
\multicolumn{1}{l}{\textbf{Case 1} \hfill
    \textit{Both Win (Baseline\cmark ~/ CMC\cmark)}} \\
\midrule
\textbf{Question:} What is the usual form of oxygen
bound compounds? \\
\textbf{Gold answer:} \textcolor{blue}{oxides} \\
\textbf{Baseline:} \textcolor{green}{oxides} \hfill EM~$=$~1 /
    F1~$=$~1.00 \\
\textbf{CMC:} \textcolor{green}{oxides} \hfill EM~$=$~1 /
    F1~$=$~1.00 \\
\midrule
\multicolumn{1}{p{14cm}}{%
\textit{Context:} Due to its electronegativity, oxygen forms chemical bonds with almost all other elements to give corresponding oxides. The surface of most metals, such as aluminium and titanium, are oxidized in the presence of air and become coated with a thin film of oxide that passivates the metal and slows further corrosion. Many oxides of the transition metals are non-stoichiometric compounds, with slightly less metal than the chemical formula would show. For example, the mineral FeO (wüstite) is written as Fe
1 - xO, where x is usually around 0.05.} \\

%% ── Case 2: CMC wins ─────────────────────────────────────
\toprule
\multicolumn{1}{l}{\textbf{Case 2} \hfill
    \textit{CMC wins (Baseline\xmark ~/ CMC\cmark)}} \\
\midrule
\textbf{Question:} What is the estimated cost of
the V\&A branded gallery? \\
\textbf{Gold answer:} \textcolor{blue}{£76 million} \\
\textbf{Baseline:} \textcolor{red}{£43 million} \hfill
    EM~$=$~0 / F1~$=$~0.50 \\
\textbf{CMC:} \textcolor{green}{£76 million} \hfill EM~$=$~1 /
    F1~$=$~1.00 \\
\midrule
\multicolumn{1}{p{14cm}}{%
\textit{Context:}  The V\&A is in discussion with the University of Dundee, University of Abertay, Dundee City Council and the Scottish Government with a view to opening a new £43 million gallery in Dundee that would use the V\&A brand although it would be funded through and operated independently. As of 2015, with costs estimated at £76 million, it is the most expensive gallery project ever undertaken in Scotland. The V\&A Dundee will be on the city's waterfront and is intended to focus on fashion, architecture, product design, graphic arts and photography. It is planned that it could open within five years. Dundee City Council is expected to pay a major part of the running costs. The V\&A is not contributing financially, but will be providing expertise, loans and exhibitions.} \\

%% ── Case 3: Baseline wins ────────────────────────────────
\toprule
\multicolumn{1}{l}{\textbf{Case 3} \hfill
    \textit{Baseline wins (Baseline\cmark ~/ CMC\xmark)}} \\
\midrule
\textbf{Question:} When was 7 Lincoln Square
completed? \\
\textbf{Gold answer:} \textcolor{blue}{June 1979} \\
\textbf{Baseline:} \textcolor{green}{June 1979} \hfill
    EM~$=$~1 / F1~$=$~1.00 \\
\textbf{CMC:} \textcolor{red}{1979} \hfill EM~$=$~0 /
    F1~$=$~0.67 \\
\midrule
\multicolumn{1}{p{14cm}}{%
\textit{Context:} Meanwhile, ABC News, which formed as a newly separate division, sought to become a global leader in television news. In 1977, Roone Arledge was named president of the new ABC News in addition to being president of ABC Sports. That same year, ABC launched a major expansion of its office facilities in New York City. The company first constructed a new 10-story building on land previously occupied by an abandoned warehouse on the corner of Columbus Avenue and West 66th Street; the facility that was built in its place is nicknamed "7 Lincoln Square" (although it is actually located at 149 Columbus Avenue). Meanwhile, a former parking lot, located at 30 West 67th Street, was transformed into an impressive 15-story building. Both buildings were completed in June 1979. WABC-TV moved its operations from offices at 77 West 66th Street to 149 Columbus Avenue, freeing up space for the ABC network to house some of its operations.} \\

%% ── Case 4: CMC partial gain ─────────────────────────────
\toprule
\multicolumn{1}{l}{\textbf{Case 4} \hfill
    \textit{CMC partial gain (Baseline$\sim$ ~/ CMC$\sim$)}} \\
\midrule
\textbf{Question:} Where did the residents of
Antioch flee to? \\
\textbf{Gold answer:} \textcolor{blue}{The city's residents fled to the north} \\
\textbf{Baseline:} \textcolor{red}{the north} \hfill
    EM~$=$~0 / F1~$=$~0.40 \\
\textbf{CMC:} \textcolor{green}{to the north} \hfill EM~$=$~0 /
    F1~$=$~0.55 \\
\midrule
\multicolumn{1}{p{14cm}}{%
\textit{Context:} The plague struck various countries in the Middle East during the pandemic, leading to serious depopulation and permanent change in both economic and social structures. As it spread to western Europe, the disease entered the region from southern Russia also. By autumn 1347, the plague reached Alexandria in Egypt, probably through the port's trade with Constantinople, and ports on the Black Sea. During 1347, the disease travelled eastward to Gaza, and north along the eastern coast to cities in Lebanon, Syria and Palestine, including Ashkelon, Acre, Jerusalem, Sidon, Damascus, Homs, and Aleppo. In 1348–49, the disease reached Antioch. The city's residents fled to the north, most of them dying during the journey, but the infection had been spread to the people of Asia Minor.} \\
\bottomrule
\end{tabular}
\caption{Qualitative case study on SQuAD validation
set (full-data setting, GPT2-Large, $r=4$, Llama).
\textit{Baseline} receives the local
context window without compression. Gold answers
are shown in \textcolor{blue}{blue}, correct
predictions in \textcolor{green}{green}, and
incorrect predictions in \textcolor{red}{red}.
EM and F1 are per-example scores.}
\label{tab:case_study}
\end{table*}

\end{document}